\PassOptionsToPackage{final}{graphicx}
\PassOptionsToPackage{final}{listings}

\RequirePackage{tikz}
\RequirePackage{pgfplots}
\documentclass[default,iicol]{sn-jnl}
\usepackage[title]{appendix}

\usepackage{hyperref}       % hyperlinks
\usepackage{url}            % simple URL typesetting
\usepackage{amsmath,amssymb,amsfonts}       % blackboard math symbols
\usepackage{booktabs}       % professional-quality tables
\usepackage{xcolor}
\usepackage{xspace}
\usepackage[disable]{todonotes}
\usepackage{paralist}
\usepackage[capitalize]{cleveref}
\usepackage{datetime2}
\usepackage{subcaption}
\usepackage{cases}
\usepackage{pgfplots}
\pgfplotsset{compat=newest}
\usepackage{wrapfig}
\usetikzlibrary{fit}

\usepackage{geometry}
\usepackage{multirow,tabularx}
\usepackage{graphicx}

\newcommand\shieldsym[1][]{%
\raisebox{-1.5pt}{\tikzset{
    shield width/.store in=\shieldwidth,
    shield width=1.5ex,
    shield height/.store in=\shieldheight,
    shield height=1.75ex
}%
\tikz [baseline,#1] \draw (0,\shieldheight) -- (0,\shieldwidth/2) arc [radius=\shieldwidth/2, start angle=-180, end angle=0] -- (\shieldwidth,\shieldheight) -- cycle;%
}}

\newcommand{\Task}{\ensuremath{\textsl{Task}}}

\newcommand{\mcresmin}[2]{\ensuremath{\eta^{\min}_{#1,#2}}}
\newcommand{\mcresmax}[2]{\ensuremath{\eta^{\max}_{#1,#2}}}

\newcommand{\val}[1]{\ensuremath{\textsl{val}_{#1}}}
\newcommand{\optval}[1]{\ensuremath{\textsl{optval}_{#1}}}

\newcommand{\shield}[2]{\ensuremath{\textsl{shield\,}_{#1}^{#2}}}
\newcommand{\shielded}[1]{\ensuremath{#1_{\text{\shieldsym}}}}
\newcommand{\online}[1]{\shielded{#1}}

\newcommand{\snake}{\textsc{Snake}\xspace}

\newcommand{\position}[1]{\textit{pos}(#1)}
\newcommand{\taskq}[1]{\textit{task}(#1)}
\renewcommand{\turn}[1]{\textit{turn}(#1)}
\newcommand{\avatar}{\textit{ava}}

\newcommand{\Distr}{\mathit{Distr}}

\newcommand{\distDom}{X}

\newcommand{\distFunc}{\mu}
\newcommand{\distDomElem}{x}
\newcommand{\supp}{\mathit{supp}}

\def\AVGREW{avg. rew.}

\newcommand{\finally}{\lozenge}
\newcommand{\sinit}{s_{\mathit{I}}} % initial state of DTMC/MDP
\newcommand{\states}[1][]{\mathcal{S}_{#1}}

\newcommand{\dtmc}{\mathcal{D}}
\newcommand{\DtmcInit}[1][]{\ensuremath{\dtmc{#1}=(\mathcal{S}{#1},\sinit{#1},P{#1})}}

\newcommand{\mdp}{\mathcal{M}}

\newcommand{\MdpTupleR}[1][]{\ensuremath{(\mathcal{S}{#1}, s_0, \Act{#1},\pmdp{#1})}}
\newcommand{\MdpInitR}[1][]{\ensuremath{\mdp{#1}=\MdpTupleR[#1]}}

\newcommand{\act}[1][a]{\alpha} % single action of MDP
\newcommand{\Act}{\mathcal{A}}        % action set of MDP

\newcommand{\rewFunction}{\ensuremath{{r}}}

\newcommand{\pmdp}{\mathcal{P}}

\newcommand{\ltp}[1][L]{\mathcal{L}}   % linear-time property
\newcommand{\bddstate}[1]{s_{\mathbb{B}}{}}

\newcommand{\tool}[1]{\texttt{#1}\xspace}

\newcommand{\prism}{\tool{PRISM}}
\newcommand{\storm}{\tool{Storm}}
\newcommand{\pytorch}{\tool{PyTorch}}

\DeclareRobustCommand{\cpp}
{\valign{\vfil\hbox{##}\vfil\cr
   \textsf{C\kern-.1em}\cr
   $\hbox{\fontsize{\sf@size}{0}\textbf{+\kern-0.05em+}}$\cr}%
}

\selectcolormodel{cmy}

\definecolor{lgrey}{rgb}{0.8,0.8,0.8}
\definecolor{grey}{rgb}{0.5,0.5,0.5}

\definecolor{lightblue}{rgb}{0.8,0.8,1.0}
\definecolor{lightred}{rgb}{1.0,0.8,0.8}
\definecolor{lightgreen}{rgb}{0.8,1.0,0.8}
\definecolor{angrygreen}{cmyk}{0.279,0,0.91,0.08}
\definecolor{lightred}{rgb}{1.0,0.8,0.8}
\definecolor{pink}{rgb}{1.0,0.1,1.0}

\definecolor{prismgreen}{HTML}{009900}
\definecolor{prismred}{HTML}{cc0000}
\definecolor{prismblue}{HTML}{0000cc}

\normalbaroutside
\newtheorem{example}{Example}
\newtheorem{definition}{Definition}

\newcommand{\response}[1]{\textcolor{black}{#1}}

\begin{document}
%\title{Online Shielding for Stochastic Systems}
\title[Online Shield RL]{Online Shielding for Reinforcement Learning}

\author*[1,4]{\fnm{Bettina} \sur{K\"{o}nighofer}}\email{bettina.koenighofer@iaik.tugraz.at}
\author[1]{\fnm{Julian} \sur{Rudolf}}\email{julian.rudolf@student.tugraz.at}
\author[1]{\fnm{Alexander} \sur{Palmisano}}\email{alexander.palmisano@student.tugraz.at}
\author[2,3]{\fnm{Martin} \sur{Tappler}}\email{martin.tappler@ist.tugraz.at}
\author[1]{\fnm{Roderick} \sur{Bloem}}\email{roderick.bloem@iaik.tugraz.at}

\affil[1]{\orgdiv{Institute of Applied Information Processing and Communications}, \orgname{Graz University of Technology}, \orgaddress{\city{Graz}, \country{Austria}}}
\affil[2]{\orgdiv{Institute of Software Technology}, \orgname{Graz University of Technology}, \orgaddress{\city{Graz}, \country{Austria}}}
\affil[3]{\orgdiv{TU Graz-SAL DES Lab}, \orgname{Silicon Austria Labs}, \orgaddress{\city{Graz}, \country{Austria}}}
\affil[4]{\orgdiv{Lamarr Security Research}, \orgaddress{\city{Graz}, \country{Austria}}}

\abstract{
	\response{Besides the recent impressive results %and the growing application areas 
on reinforcement learning (RL), safety
%especially during exploration 
is still one of the major research challenges in RL.} 
\response{RL is a machine-learning approach to determine near-optimal policies in Markov decision processes (MDPs).
In this paper, we consider the setting where the safety-relevant fragment of the MDP
together with a temporal logic safety specification is given and many safety violations can be avoided by planning ahead a short time into the future.}
\response{We propose an approach for online safety shielding of RL agents. During runtime, 
the shield analyses the safety of each available action.
For any action, the shield computes the maximal probability to not violate the safety specification within 
the next $k$ steps when executing this action. 
Based on this probability and a given threshold, the shield decides whether to block an action from the agent.
}
Existing offline shielding approaches compute exhaustively the safety of all state-action combinations ahead of time, resulting in huge computation times and large memory consumption.%, and significant delays at runtime due to the look-ups in huge databases. 
The intuition behind online shielding is to compute at runtime the set of all states that could be reached in the near future. For each of these states, the safety of all available actions is analysed and used for shielding as soon as one of the considered states is reached. %Empirically, we study the effects on the learning performance and the safety of the learned policy when penalizing blocked actions during training. 
\response{Our approach is well suited for high-level planning problems where the time between decisions can be used for safety computations and it is sustainable for the agent to wait until these computations are finished}.
For our evaluation, we selected a 2-player version of the classical computer game \snake. The game \response{represents a high-level planning problem} that requires fast decisions and the multiplayer setting induces a large state space, which is computationally expensive to analyse exhaustively. %\response{The safety objective
}

\maketitle

\section{Introduction}

%\paragraph{Motivation - Safe RL.}
Reinforcement Learning (RL) has proven successful in solving complex tasks that are difficult to solve using classic controller design, including applications in computer games~\cite{silver2016mastering}, multi-agent planning~\cite{DBLP:journals/corr/abs-1910-12639}, and robotics~\cite{wang2019learning}. 
RL learns high-performance controllers by optimising objectives expressed via rewards in unknown, stochastic environments. 
Although learning-enabled controllers (LECs) have the potential to outperform classical controllers, safety concerns prevent LECs from being widely used in real-world tasks~\cite{amodei2016concrete}.

\response{\emph{Shielding}~\cite{DBLP:conf/tacas/BloemKKW15}
 is a runtime enforcement technique to ensure safe decision making. 
By augmenting an RL agent with a \emph{shield}, at every time step, 
unsafe actions 
are blocked by the shield and the learning agent can only pick a 
safe action to be sent to the environment. Shields are automatically constructed via correct-by-construction formal synthesis methods from a model of the safety-relevant environment dynamics and a safety specification.
In the deterministic setting, shields ensure that unsafe states specified by the safety specification are
\emph{never} visited. Consequently, in the absence of uncertainties, an agent augmented with a shield 
is guaranteed to satisfy the safety objective as long as the shield is used.}

\response{In scenarios that incorporate uncertainties, probabilistic shields have been used to 
reduce safety violations during training and execution~\cite{DBLP:conf/concur/0001KJSB20}.
The premise of this work is that often in real-world applications,
many safety violations can be avoided by analysing the consequences of actions in \emph{the near future}.
To compute probabilistic shields, the safety-relevant dynamics of
the environment are modelled as a Markov Decision Process (MDP)
and the specification $\varphi$ is expressed in a safety fragment of linear temporal logic~\cite{BK08}.
Such a specification could, for example, forbid to reach a set of critical states in the MDP.}

\response{For each state and action, exact probabilities are computed on how likely it is that executing the action results in violating $\varphi$ from the current state within the next $k$ steps.
At a state $s$, an actions  $a$ is called \emph{unsafe} if executing $a$ incurs a probability of violating $\varphi$ within the next $k$ steps greater than a threshold $\delta$ with respect to the optimal safety probability possible in $s$.
During runtime, the probabilistic shield then blocks any unsafe action from the agent.
A probabilistic shield can be used to shield the decisions of an agent in the training as well as in the execution phase. 
In this paper, we build on the approach of~\cite{DBLP:conf/concur/0001KJSB20}. Hence, from here on,
we simply refer to probabilistic shields as shields.}

%In RL, optimal strategies are obtained without prior knowledge about the environment. Therefore, the safety of actions is not known before their executions. Even after training, there is no guarantee that no unsafe actions are part of the final policy. Having no safety guarantees is unacceptable for safety-critical areas, such as autonomous driving. These guarantees take different forms. Especially safety-critical operations require the absence of all unsafe behaviour, while achieving absolute safety for all operations may be impossible due to uncertain, stochastic behaviour. In these cases, safety guarantees may limit the probability of unsafe events.
%@COMMENT of reviewers that the tern "Safety guarantee" implies absolute safety, which we cannot guarantee.  

\emph{The problem with offline shielding.}
The computation of an offline shield for discrete-event systems requires an exhaustive, ahead-of-time safety analysis for all possible state-action combinations.
%COMMENT of reviewer 2 regarding continuous stuff
% added "for discrete-event systems"
Therefore, the complexity of offline shield synthesis grows exponentially in the state and action dimensions, which limits the application of offline shielding to small environments.
Previous work that applied shields in complex, high-dimensional environments
relied on over-approximations of the reachable states and domain-oriented abstractions~\cite{shield_rl,DBLP:conf/cav/AvniBCHKP19}. However, this may result in imprecise safety computations of the shield. This way, the shield may become over-restrictive, hindering the learning agent in properly exploring the environment and finding its optimal 
policy~\cite{DBLP:conf/concur/0001KJSB20}.

\emph{Our solution -- online shielding.} 
Our approach is based on the idea of computing the safety of actions on-the-fly during run time.
In many applications, the learning agent does not have to take a decision at every time step.
Instead, the learning agent only has to make a decision when reaching a \emph{decision state}.
As an example consider a service robot %that is controlled to traverse 
traversing a corridor. 
The agent has time until the service robot reaches the end of the corridor, i.e., the next decision state, to decide where the service robot should go next. %UBER, better example, replace me later!
Online shielding uses the time between two decision states to
compute the safety of all possible actions in the next decision state. When reaching the next
decision state, this information is used to block unsafe actions of the agent.
While the online safety analysis incurs a runtime overhead, every single computation of the safety of an action is efficient and parallelisable. 
Thus, in many settings, expensive offline pre-computation and huge shielding databases with costly lookups are not necessary. 
Since the safety analysis is performed only for decision states that are actually reached,
online shielding is applicable to large, changing, or unknown environments.

We address the problem of shielding a controllable RL agent in an environment shared with other autonomous agents that perform tasks concurrently. For example, some combinations of agent positions may be unsafe, as they correspond to collisions. \response{The specification then forbids visiting such states, i.e., $\varphi = \mathbf{G}(\lnot \mathcal{S}_{collision})$.} %The task of the shield is to block unsafe actions, i.e., actions that increase the probability of violating $\varphi$ withing the next $k$ steps by more than a threshold $\delta$ with respect to the optimal probability.}

In online shielding, the computation of the safety for any action in the next decision state starts as soon as the controllable agent leaves the current decision state.
The tricky part of online shielding in the multi-agent setting is that during the time the RL agent has between two consecutive decisions,
the other agents also change their positions. Therefore, online shielding requires to compute the safety of actions with respect to all possible movements of the other agents.
\response{As soon as the next decision state is reached, the results from the safety analysis are used to block unsafe actions.} 

Technically, we use MDPs to formalise the dynamics of the agents operating within the environment.
At runtime, we create a sub-MDP for each action.
These sub-MDPs model the immediate future from the viewpoint of the RL agent. Via model checking, we determine for the next \response{$k$ steps} the minimal probability of violating $\varphi$.
An action unsafe and therefore blocked by the shield, if the action violates safety with a probability higher than a threshold $\delta$. We generally set $\delta$ relative to the minimal probability of violating \response{$\varphi$ within the next $k$ steps. In some state, any available action may impose 
a large risk of violating $\varphi$, while in other states there may exists actions that guarantee to stay safe within the next $k$ steps. Using a relative threshold allows an \emph{adaptive} notion of shielding and ensures deadlock freedom by always allowing at least one action.}

\response{\emph{Requirements and Limitations.}}
\response{The online shielding approach relies upon some requirements and is affected
by a few limitations that we want to briefly summarise. 
The first limitation addresses the fact that the proposed approach does not provide
any worst-case computation times. Therefore the possibility that the agent reaches the next decision state before the shield made its decision on which action to block cannot be ruled out. 
Online shielding is therefore only applicable in settings that allow an alternative course of action such as "waiting" if the safety analysis is not completed.
In general, we recommend to use online shielding in settings where the average time
between visiting decision states is larger than the average time used for the safety analysis.}

\response{As a second limitation we list the possible state-space explosion of the individual sub-MDPs. The size of the sub-MDPs depends among others on the used finite horizon $k$,
the number of agents operating within the environment and the number of decision states within the next $k$ steps. In extreme 
cases where every state is a decision state, online shielding would likely
not be applicable, because (1) there would be almost no time between
the individual decisions and (2) the state space of the constructed sub-MDPs would
explode.}

\response{Third, the safety of actions is only analysed within a finite horizon $k$.
Therefore, the agent might end up in situations where any available action induces a high probability of violating the specification. It is therefore important to pick a finite horizon $k$ large enough to prevent such situations. The minimal size of $k$ needed to
prevent many safety violations depends on the concrete setting and there is a natural trade-off between the computational overhead for the safety-analysis and the number of safety violations that can be prevented by the shield. As a reference value, we recommend to use a finite horizon $k$ larger than the number of steps between any two adjacent decision states.}

%\response{
%One of the main requirements, which we assume to hold, is that the time 
%between decisions is long enough to facilitate shield computations.
%Closely related to that, the ratio between decision locations and 
%non-decision locations needs to be low enough. In extreme 
%cases where every location is a decision location, online shielding would likely
%not be applicable. First, because there would be almost no time between
%individual decisions. Second, the concrete state space of constructed sub-MDPs would
%explode.
%A related limitation is that we currently do not provide worst-case
%computation times, thus online shielding does not satisfy strict 
%real-time requirements. 
%}

%\response{With respect to safety, we provide guarantees only within the considered finite horizon.
%However, as we compute new shields online, we ensure that the horizon
%is the same at every new decision location.
%}

\emph{Contributions.} 
This paper is an extended version of~\cite{DBLP:conf/nfm/KonighoferRPTB21}, in which we gave the formalisation of online shielding in probabilistic environments and presented experimental results of shielding a simple tabular Q-learning agent for a 2-player version of the classical  computer  game \snake\footnote{\url{http://www.onlineshielding.at},  \DTMdisplaydate{2020}{11}{27}{-1}}.
The evaluation demonstrated that shields can be efficiently computed at runtime, guarantee safety, and have the potential to positively influence learning performance.

The novelty of this work with respect to~\cite{DBLP:conf/nfm/KonighoferRPTB21} is an extensive case study that focuses on the question:
Can we use online shielding to \emph{safely learn a safe policy?}

An unshielded agent learns about the safety of actions by exploring actions and receiving negative rewards if an action is unsafe.
In case of standard shielded learning, all unsafe actions are blocked from the agent and the agent never gets the chance to learn the safety constraints from the negative rewards.
Since an unsafe action may look like a safe alternative to the RL agent due to the absence of negative rewards, unsafe actions might very likely be part of the final policy.
To learn a safe policy under shielding, we propose to apply \emph{informed shields}. These shields update the value function of the agent with negative rewards for any action that is blocked by the shield, using different thresholds for the comparison of the safety values. By investigating unshielded learning and learning with informed and uninformed shields in our case study, we empirically assess
how shields affect convergence to an optimal and safe policy. Since learning a safe policy is only possible if the reward structure captures
all relevant safety constraints~\cite{DBLP:conf/hybrid/HuntFMHDS21}, uninformed shields that merely block actions may hinder learning a safe policies.

%---> Wait when not done
%For each of these states, the safety
%of all available actions is analysed and used for shielding as soon as one of the considered states is
%reached. Our approach is well suited for high-level planning problems where the time between deci-
%sions can be used for safety computations and it is sustainable for the agent to wait until these
%computations are finished.

% to hold, is
%to facilitate shield computations. Closely related
%to that, the ratio between decision locations and
%non-decision locations needs to be low enough.
%In extreme cases where every location is a deci-
%sion location, online shielding would likely not be
%applicable. First, because there would be almost
%no time between individual decisions. Second, the
%oncrete state space of constructed sub-MDPs
%ould explode. A related limitation is that we
%currently do not provide worst-case computation
%times, thus online shielding does not satisfy strict
%real-time requirements.

We performed the case study on a Deep Q-learning agent for 2-Player \snake to study the safety of the final policies in various settings. 
We compared the final policies of unshielded agents and agents augmented with informed and uninformed shields. %and without policy updates with negative rewards for blocked actions.
The results show that shielding leads to a better performance during learning in all cases.
% However, in most experiments, the final policies of shielded agents inflict more unsafe behaviour than the policies of unshielded agents. even though blocked actions where updated with negative rewards.
However, in most experiments, the final policies obtained by shielded learning  produce more unsafe behaviour than the policies learned by unshielded agents. This happens with both uninformed and informed shields that provide negative rewards for blocked actions.
These results suggest that in shielded learning, the shield is also needed in the execution phase.

\emph{Outline.}
The rest of the paper is structured as follows. \cref{sec:related_work} discusses related work. We discuss the relevant foundations
in Section~\ref{sec:preliminaries}. 
In Section~\ref{sec:problem}, we present the setting and formulate the problem that we address. We present online shielding in Section~\ref{sec:shields}, by defining semantics for autonomous agents in the considered setting and defining online shield computations based on these semantics. In Section~\ref{sec:experiments}, we report on the evaluation of online shielding for the classic computer game \snake. \cref{sec:conclusion} concludes the paper with a summary and an outlook on future work.

\subsection{Related Work}
\label{sec:related_work}

\response{In reinforcement learning (RL)~\cite{sutton1998reinforcement}, an
agent aims to compute an optimal policy that maximizes the expected total amount of reward through trial-and-error via interactions with an unknown environment. While exploring unknown state-action pairs, RL agents that are 
agnostic %with respect 
to safety may undeliberately execute unsafe actions.
%without deliberateness. 
Safe RL algorithms aim to guarantee safety even during exploration, at least with high probability.}

\response{Previous approaches to safe RL include teacher advice~\cite{garcia2015comprehensive}, reward-shaping~\cite{HasanbeigAK20,DBLP:journals/corr/abs-1902-00778,DBLP:conf/tacas/HahnPSSTW19}, and policy optimisation using constraints~\cite{DBLP:conf/hybrid/HuntFMHDS21,DBLP:conf/atal/GiacobbeHKW21}.}

Several  recent  works~\cite{DBLP:journals/corr/abs-1902-00778,DBLP:conf/tacas/HahnPSSTW19} considered logically-constrained  RL, which  employs  temporal  logic as  formal  reward  shaping  technique.
The final policy of an agent trained with such a reward structure maximises the probability of satisfying the specified formula. In the context of safe RL, the formula can express a safety property and the trained agent will minimise the risk of violating the property. 
In order to ensure safety during training, logically-constrained RL has to be extended by restricting the exploration during training ~\cite{HasanbeigAK20}.

\response{We follow a model-based approach for safe RL.
In the continuous domain, model-based approaches have utilised Lyapunov-based
methods~\cite{chow2018lyapunov} and control-barrier functions~\cite{cheng2019end,ohnishi2018safety} to  enable safe learning under continuous system dynamics.
Li et al.~\cite{DBLP:conf/icra/LiB20} proposed model predictive shielding (MPS) for continuous systems.
Given an optimal policy and a safe policy, MPS checks online before executing 
the next action of the optimal policy whether the new state would allow 
reaching an invariant safe state within the next $k$ steps when executing the safe policy.
If not, MPS does not execute the action of the optimal policy and switches to the safe policy instead.
In the control community, this architecture is also known as the simplex architecture and gave rise to several recent applications in RL~\cite{ionescu2021adaptive}.
In this architecture, a switching logic switches from an advanced controller (the RL agent) to a verified base controller as soon as a state would be visited that is outside of the region where the base controller is guaranteed to satisfy an invariant safety property.
In the general shielding setting, shielding supports safety specifications beyond invariant properties, including, for example, bounded liveness properties. Furthermore, compared to MPS and the simplex architecture, shielding interferes less with the agent. Instead of overwriting an unsafe action chosen by the agent with a particular safe one, a shield lets the agent choose any action as long as it is safe}.

\response{Fulton et al.~\cite{DBLP:conf/aaai/FultonP18} published the first work
on verifiable safe learning for hybrid systems. In this work, a theorem prover is used to prove the correctness of a model
with respect to a safety specification given in differential dynamic logic.
This work was extended in~\cite{DBLP:conf/tacas/FultonP19} to the setting in which a single accurate model is not known at design time. 
The authors propose an approach in which multiple environmental models that are provably correct are constructed at design time.
During runtime, based on the collected data, the approach selects between the available models.
A major drawback of the approaches from~\cite{DBLP:conf/aaai/FultonP18,DBLP:conf/tacas/FultonP19} is that they only
learn control policies over handcrafted symbolic state spaces.
This limitation was addressed in~\cite{DBLP:conf/hybrid/HuntFMHDS21}.
Prior to RL, an agent is trained to detect positions of safety-critical objects from visual data.
During RL, this information is then used to enforce formal safety constraints
that take noise from the object detection systems into account.}

\response{In the discrete domain, the shielding approach is commonly used in safe RL~\cite{shield_rl,DBLP:conf/isola/KonighoferL0B20}.
In the deterministic setting, shields are usually constructed offline by computing a maximally permissive policy containing all actions that will not violate the safety specification on the infinite horizon. Jansen et al.~\cite{DBLP:conf/concur/0001KJSB20} introduced offline shielding in probabilistic environments, considering safety within a finite horizon.
Giacobbe et al.~\cite{DBLP:conf/atal/GiacobbeHKW21} used a very similar approach to shield safety properties in Atari games.
Our work directly extends the approach to shielding by Jansen et al.~\cite{DBLP:conf/concur/0001KJSB20} to the online setting.
Several resent extensions of shielding in probabilistic environments exist, like shielding under partial observability~\cite{DBLP:journals/corr/abs-2204-00755},
shielding quantitative properties~\cite{DBLP:journals/corr/abs-2010-03842}, or 
shielding multi-agent systems~\cite{DBLP:conf/atal/Elsayed-AlyBAET21}.
In 2021, the shield synthesis tool TEMPEST was published, which is able to synthesise several different notions of shields proposed in literature for probabilistic environments~\cite{DBLP:conf/atva/PrangerKPB21}.
}

\response{\emph{Novelty of our work.}
All  model-based safe RL approaches discussed in this section rely on building the model (or several models) at design time
and exhaustively analysing the safety of actions within the model at design time.
Our approach builds the environmental models for a finite horizon at runtime
and performs the safety verification online.
This allows our approach to be applied on large, at design time unknown, and changing environments.}

\response{To make the safety verification at runtime possible, we are the first to compute the safety of actions before a decision state is visited, by considering all possible movements of adversarial agents in the computations.
By analysing the actions before a decision state is reached, we prevent delays at runtime.}

\section{Preliminaries}
\label{sec:preliminaries}
%In this section, we introduce models and properties considered in this paper.

\noindent
\emph{Sequence and Tuple Notation.}
We denote sequences of elements by $t = e_0 \cdots e_n$ with $\epsilon$ denoting the empty sequence. The length of $t$ is denoted as $|t| = n+1$. We use  $t[i] = e_i$ for $0$-based indexed access on tuples and sequences. The notation $t[i \gets e_i']$ represents overwriting of the $i$\textsuperscript{th} element of $t$ by $e_i'$, that is, $t[j] = t[i \gets e_i'][j]$ for all $j\neq i$ and $t[i \gets e_i'][i] = e_i'$. 

A \emph{probability distribution} over a countable set $\distDom$ is a function $\distFunc\colon\distDom\rightarrow[0,1]$ with $\sum_{\distDomElem\in\distDom}\distFunc(\distDomElem)=1$.
$\Distr(\distDom)$ denotes all distributions on $\distDom$. The support of $\distFunc\in\Distr(\distDom)$ is $\supp(\distFunc)=\{x\in\distDom \mid \distFunc(x){>}0\}$.

A \emph{Markov decision process} (MDP) $\MdpInitR$ is a tuple
with a finite set $\states$ of states, a unique initial state $s_0 \in \states$, a finite set $\Act=\{a_1\dots, a_n\}$ of actions, and
a (partial) \emph{probabilistic transition function} $\pmdp: \states \times \Act \rightarrow Distr (\states)$, where $\pmdp(s,a) = \bot$ denotes undefined behaviour.
For all $s \in \states$ the available actions are $\Act(s) = \{a \in \Act | \pmdp(s, a) \neq \bot\}$ and we assume $|\Act(s)| \geq 1$.
A \emph{path} in an MDP $\mdp$ is a finite (or infinite) sequence $\rho=s_0a_0s_1a_1\ldots$ with $\pmdp(s_i, a_i)(s_{i+1}) > 0$ for all $i\geq 0$ 
unless otherwise noted.

Non-deterministic choices in an MDP are resolved by a
so-called \emph{policy}. 
For the properties considered in this paper, memoryless deterministic policies are sufficient~\cite{BK08}. These are functions $\pi : \states \rightarrow \Act$ with $\pi(s) \in \Act(s)$.
We denote the set of all memoryless deterministic policies of an MDP by $\Pi$.
Applying a policy $\pi$ to an MDP yields an induced \emph{Markov chain}~ \DtmcInit ~with $P : \mathcal{S} \rightarrow\Distr(\states)$ where all nondeterminism is resolved.
A \emph{reward function} $\rewFunction \colon \mathcal{S} \times \Act \rightarrow \mathbb{R}$ for an MDP adds a reward  to every state $s$ and action $a$. %enabled in $s$.

In formal methods, safety properties are often specified as \emph{linear temporal logic} (LTL) formulas~\cite{pnueli1977temporal}.
For an MDP $\mdp$, probabilistic model checking~\cite{Kat16,DBLP:conf/lics/Kwiatkowska03}
 employs value iteration or linear programming to compute the probabilities of \emph{all states and actions of the MDP}
 to satisfy a safety property $\varphi$.

Specifically, we compute $\mcresmax{\varphi}{\mdp} \colon \mathcal{S} \rightarrow [0,1]$ or $\mcresmin{\varphi}{\mdp} \colon \mathcal{S} \rightarrow [0,1]$,
which yields for all states the maximal (or minimal) probability over all possible policies to satisfy $\varphi$.
For instance,  for $\varphi$ encoding to reach a set of states $T$,  $\mcresmax{\varphi}{\mdp}(s)$ is the maximal probability to ``eventually'' reach a state in $T$
from state $s\in \mathcal{S}$.

\section{Setting and Problem Statement}\label{sec:problem}
\paragraph{Setting}
We consider a setting similar to~\cite{DBLP:conf/concur/0001KJSB20}, 
where one controllable agent, called the \emph{avatar}, and multiple uncontrollable
agents, called \emph{adversaries} operate within an \emph{arena}.
The arena is a compact, high-level
description of the underlying model and captures the dynamics of the agents. 
Any information on rewards is neglected within the arena since it is not needed for safety computations. 

From this arena, potential agent locations may be inferred. 
Within the arena, the agents perform \emph{tasks} that are sequences of \emph{activities} performed consecutively.

% \begin{example}[Robot logistics in a smart factory]
% Take a factory floor plan with several corridors with machines.
% The adversaries are (possibly autonomous) transporters moving parts within the factory.
% The avatar models a specific service unit moving around and inspecting machines where an issue has been raised, while accounting for the behaviour of the adversaries.
%  Corridors might be too narrow for multiple robots, which poses a safety-critical situation.
% A task of the avatar may correspond to inspecting machines. This task consists of several activities that define the path that needs to be taken to visit the machine. 
% \end{example}

Formally, an \emph{arena} is a pair $G = (V, E)$, where $V$ is a set of nodes and $E$ is a finite set of  $E \subseteq V\times V$.
An agent’s \emph{location} is defined via the current node $v \in V$.
An edge $(v,v')\in E$ represents an \emph{activity} of an agent. By executing an activity, the agent moves to its next location $v'$.
A \emph{task} is defined as a non-empty sequence $(v_1,v_2) \cdot (v_2,v_3) \cdot (v_3, v_4) \cdots (v_{n-1},v_n) \in E^*$ of connected edges. 
To ease representation, we denote tasks also as sequences of locations $v_1 \cdot v_2 \cdots v_n$. 

The set of tasks available in a location $v \in V$ is given by the function $\Task(v)$.
The set of all tasks of an arena $G$ is denoted by $\Task(G)$.
The avatar is only able to select a next task at a \emph{decision location} in $V_{D}\subseteq V$.
To avoid deadlocks, we require for any decision location $v\in V_{D}$ that $\Task(v) \neq \emptyset$ and for all $v \cdots v' \in Task(v)$ that $v' \in  V_{D}$, i.e., any task ends in another decision location from which the agent is 
able to decide on a new task.
A safety property may describe that some combinations of agent positions are unsafe %safety-critical
and should not be reached
(or any other safety property from the safety fragment of LTL).

%\begin{wrapfig}[t]
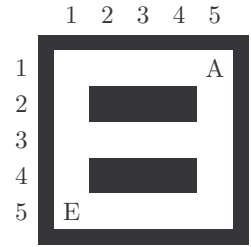
\begin{figure}[t]
\begin{center}
\begin{tikzpicture}[scale=0.95, transform shape]
\draw[step=0.5cm,color=gray] (-1,-1) grid (1.5,1.5);

\node[minimum width = 0.5cm, minimum height = 0.2cm, inner sep = 0pt, outer sep = 0pt] at ( -0.75,+2.0) {1};
\node[minimum width = 0.5cm, minimum height = 0.2cm, inner sep = 0pt, outer sep = 0pt] at ( -0.25,+2.0) {2};
\node[minimum width = 0.5cm, minimum height = 0.2cm, inner sep = 0pt, outer sep = 0pt] at (  0.25,+2.0) {3};
\node[minimum width = 0.5cm, minimum height = 0.2cm, inner sep = 0pt, outer sep = 0pt] at (  0.75,+2.0) {4};
\node[minimum width = 0.5cm, minimum height = 0.2cm, inner sep = 0pt, outer sep = 0pt] at (  1.25,+2.0) {5};

\node[minimum width = 0.2cm, minimum height = 0.5cm] at ( -1.45,1.25) {1};
\node[minimum width = 0.2cm, minimum height = 0.5cm] at ( -1.45,0.75) {2};
\node[minimum width = 0.2cm, minimum height = 0.5cm] at ( -1.45,0.25) {3};
\node[minimum width = 0.2cm, minimum height = 0.5cm] at ( -1.45,-0.25) {4};
\node[minimum width = 0.2cm, minimum height = 0.5cm] at ( -1.45,-0.75) {5};

\node[fill=black, minimum width = 0.2cm, minimum height = 0.2cm] at ( -1.1,+1.6) {};
\node[fill=black, minimum width = 0.5cm, minimum height = 0.2cm] at ( -0.75,+1.6) {};
\node[fill=black, minimum width = 0.5cm, minimum height = 0.2cm] at ( -0.25,+1.6) {};
\node[fill=black, minimum width = 0.5cm, minimum height = 0.2cm] at (  0.25,+1.6) {};
\node[fill=black, minimum width = 0.5cm, minimum height = 0.2cm] at (  0.75,+1.6) {};
\node[fill=black, minimum width = 0.5cm, minimum height = 0.2cm] at (  1.25,+1.6) {};
\node[fill=black, minimum width = 0.2cm, minimum height = 0.2cm] at (  1.6,+1.6) {};

\node[fill=black, minimum width = 0.2cm, minimum height = 0.2cm] at ( -1.1,-1.1) {};
\node[fill=black, minimum width = 0.5cm, minimum height = 0.2cm] at ( -0.75,-1.1) {};
\node[fill=black, minimum width = 0.5cm, minimum height = 0.2cm] at ( -0.25,-1.1) {};
\node[fill=black, minimum width = 0.5cm, minimum height = 0.2cm] at (  0.25,-1.1) {};
\node[fill=black, minimum width = 0.5cm, minimum height = 0.2cm] at (  0.75,-1.1) {};
\node[fill=black, minimum width = 0.5cm, minimum height = 0.2cm] at (  1.25,-1.1) {};
\node[fill=black, minimum width = 0.2cm, minimum height = 0.2cm] at (  1.6,-1.1) {};

\node[fill=black, minimum width = 0.2cm, minimum height = 0.5cm] at ( -1.1,-0.75) {};
\node[fill=black, minimum width = 0.2cm, minimum height = 0.5cm] at ( -1.1,-0.25) {};
\node[fill=black, minimum width = 0.2cm, minimum height = 0.5cm] at ( -1.1, 0.25) {};
\node[fill=black, minimum width = 0.2cm, minimum height = 0.5cm] at ( -1.1, 0.75) {};
\node[fill=black, minimum width = 0.2cm, minimum height = 0.5cm] at ( -1.1, 1.25) {};

\node[fill=black, minimum width = 0.2cm, minimum height = 0.5cm] at (  1.6,-0.75) {};
\node[fill=black, minimum width = 0.2cm, minimum height = 0.5cm] at (  1.6,-0.25) {};
\node[fill=black, minimum width = 0.2cm, minimum height = 0.5cm] at (  1.6, 0.25) {};
\node[fill=black, minimum width = 0.2cm, minimum height = 0.5cm] at (  1.6, 0.75) {};
\node[fill=black, minimum width = 0.2cm, minimum height = 0.5cm] at (  1.6, 1.25) {};

\node[fill=white, minimum width = 0.5cm, minimum height = 0.5cm] at (-0.75,+1.25) {};
%\node[minimum size=0.4cm,draw,thick] at (-0.75,+1.25) {};
\node[fill=white, minimum width = 0.5cm, minimum height = 0.5cm] at (-0.25,+1.25) { };
\node[fill=white, minimum width = 0.5cm, minimum height = 0.5cm] at ( 0.25,+1.25) { };
\node[fill=white, minimum width = 0.5cm, minimum height = 0.5cm] at ( 0.75,+1.25) { };
\node[fill=white, minimum width = 0.5cm, minimum height = 0.5cm] at ( 1.25,+1.25) {A};

\node[fill=white, minimum width = 0.5cm, minimum height = 0.5cm] at (-0.75,+0.75) {};
\node[fill=black, minimum width = 0.5cm, minimum height = 0.5cm] at (-0.25,+0.75) {};
\node[fill=black, minimum width = 0.5cm, minimum height = 0.5cm] at ( 0.25,+0.75) {};
\node[fill=black, minimum width = 0.5cm, minimum height = 0.5cm] at ( 0.75,+0.75) { };
\node[fill=white, minimum width = 0.5cm, minimum height = 0.5cm] at ( 1.25,+0.75) { };

\node[fill=white, minimum width = 0.5cm, minimum height = 0.5cm] at (-0.75,+0.25) { };
\node[fill=white, minimum width = 0.5cm, minimum height = 0.5cm] at (-0.25,+0.25) { };
\node[fill=white, minimum width = 0.5cm, minimum height = 0.5cm] at ( 0.25,+0.25) { };
\node[fill=white, minimum width = 0.5cm, minimum height = 0.5cm] at ( 0.75,+0.25) { };
\node[fill=white, minimum width = 0.5cm, minimum height = 0.5cm] at ( 1.25,+0.25) { };

\node[fill=white, minimum width = 0.5cm, minimum height = 0.5cm] at (-0.75,-0.25) { };
\node[fill=black, minimum width = 0.5cm, minimum height = 0.5cm] at (-0.25,-0.25) { };
\node[fill=black, minimum width = 0.5cm, minimum height = 0.5cm] at ( 0.25,-0.25) { };
\node[fill=black, minimum width = 0.5cm, minimum height = 0.5cm] at ( 0.75,-0.25) { };
\node[fill=white, minimum width = 0.5cm, minimum height = 0.5cm] at ( 1.25,-0.25) { };

\node[circle,minimum size=0.4cm,draw,thick] at (-0.75,-0.75) {};
\node[circle,minimum size=0.3cm,draw,thick] at (-0.75,-0.75) {};

\node[fill=white, minimum width = 0.5cm, minimum height = 0.5cm] at (-0.75,-0.75) {E};

\node[fill=white, minimum width = 0.5cm, minimum height = 0.5cm] at (-0.25,-0.75) { };
\node[fill=white, minimum width = 0.5cm, minimum height = 0.5cm] at ( 0.25,-0.75) { };
\node[fill=white, minimum width = 0.5cm, minimum height = 0.5cm] at ( 0.75,-0.75) { };
\node[fill=white, minimum width = 0.5cm, minimum height = 0.5cm] at ( 1.25,-0.75) { };
\node at (1,+1.25){};
\node at (1,-1.25){};
\end{tikzpicture}

\end{center}
\caption{Gridworld with avatar A (top right) and an adversary E (bottom left).}
\label{fig:maze}
%\end{wrapfig}
\end{figure}

\begin{example}[Gridworld]
\Cref{fig:maze} shows a simple gridworld with corridors represented by white tiles and walls represented by black tiles. 
A tile is defined via its $(x,y)$ position.
We model this gridworld with an arena $G = (V,E)$ by associating each white tile with a location in $V$ and creating an edge in $E$ for each pair of adjacent white tiles.
Corners and crossings are decision locations, i.e., $V_d = \{(1,1), (1,3), (1,5), (5,1), (5,3), (5,5)\}$. 
At each decision location, tasks define sequences of activities needed to traverse adjoining corridors, e.g., $Task((1,3)) = \{(1,3) \cdot (2,3) \cdot (3,3) \cdot (4,3) \cdot(5,3)$, 
  $(1,3) \cdot (1,2) \cdot (1,1)$,  
  $(1,3) \cdot (1,4) \cdot (1,5)\}$.
\end{example}

\paragraph{Problem Statement}%
Consider an environment described by an arena as above and a safety specification $\varphi$.
We assume stochastic behaviours for the adversaries, e.g, obtained using RL~\cite{sadigh2018planning,sadigh2016planning}.
In fact, this stochastic behaviour determines all actions of the adversaries via probabilities.
The underlying model is then an MDP: the avatar executes an action, and upon this execution, the next exact positions (the state of the system) are determined stochastically. \response{That is, the 
states correspond to the possible positions
of all agents including the avatar and
the actions correspond to the available tasks.}

\response{Our aim is to \emph{shield unsafe actions} from the avatar during training as well as during execution. At any state $s$, an actions  $a$ is called \emph{unsafe} 
if executing $a$ incurs a probability of violating $\varphi$ within the next $k$ steps greater than a threshold $\delta$ that is defined with respect to the optimal safety probability possible in $s$.}

The safety analysis of actions is performed on-the-fly allowing the avatar to operate within large arenas. 

\begin{example}[Gridworld]
In Figure~\ref{fig:maze},
the tile labelled \texttt{A} denotes the location of the avatar and the tile labelled \texttt{E} denotes the position of an adversary. 
Let $(x_A,y_A)$ and $(x_E,y_E)$ be the positions of the avatar and the adversary, respectively.
A safety property in this scenario is \response{$\varphi = \mathbf{G}(\lnot(x_A = x_E \land y_A = y_E))$. The ``globally'' operator $\mathbf{G}$ states that unsafe states must not be entered,
i.e., that collisions are never allowed. 
The shield blocks unsafe actions that 
would increase the probability of violating $\varphi$ within the next $k$ steps by more than a relative threshold $\delta$.
We give more details in Section~\ref{sec:shield_construction} on how to construct a shield for this setting.}

\end{example}

\section{Online Shielding for MDPs}
\label{sec:shields}
\begin{figure*}[tb]
\centering
\scalebox{0.85}{
\input{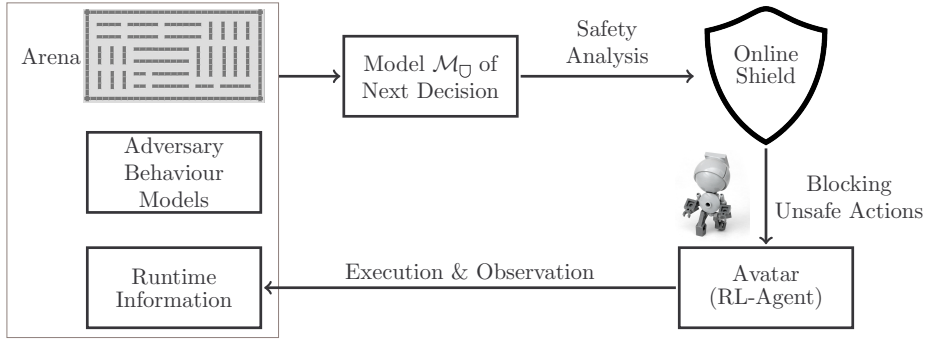}
}
\caption{Workflow of the shield construction.}
\label{fig:flowchart}
\end{figure*}

In this section, we outline the workflow of online shielding in Figure~\ref{fig:flowchart} and  describe it below. Given an arena and behaviour models for adversaries, we define an MDP $\mdp$ that captures all safety-relevant information. 
At runtime, we use current runtime information 
to create sub-MDPs $\online{\mdp}$ of $\mdp$ that model
the immediate future of the agents up to some finite horizon. 
Given such a sub-MDP $\online{\mdp}$ and a safety property $\varphi$, we 
compute via model checking the probability to violate $\varphi$ within the finite horizon
for each task available. The shield then blocks tasks involving a too large risk from the
avatar. \response{To ensure effectiveness, we choose
the horizon large enough such that it covers the
distance between any pair of adjacent decision locations, i.e., pairs of locations connected by a task.}

\subsection{Behaviour Models for Adversaries}\label{sec:learning_adv}

The adversaries and the avatar operate within a shared environment, which is represented by an arena $G=(V,E)$, and perform tasks independently.
We assume that we are given a stochastic behaviour model of each adversary that determines all task choices
of the respective adversary via probabilities.
The behaviour of an adversary is formally defined as follows.

\begin{definition}[Adversary Behaviour]
\label{def:adv_behaviour}
For an arena $G = (V,E)$, we define the behaviour $B_i$ of an adversary $i$ as a function $B_i \colon V_D \rightarrow \Distr(\Task(G))$ from decision locations to distributions over tasks, with $\supp(B_i(v)) \subseteq \Task(v)$.
\end{definition}

Behaviour models of adversaries may be derived using domain knowledge or generalised from observations using machine learning or automata learning~\cite{DBLP:journals/ml/MaoCJNLN16,DBLP:journals/fac/TapplerA0EL21,smm_learning}. A potential approach is to observe adversaries in smaller arenas and transfer knowledge gained in this way to larger arenas~\cite{DBLP:conf/concur/0001KJSB20}. Cooperative and truly adverse behaviour of adversaries may require considering additional aspects in the adversary behaviour, such as the arena state at a specific point in time. Such considerations are beyond the scope of this paper, since complex adversary behaviour generally makes the creation of behaviour models more difficult, whereas the online shield computations are hardly affected. \response{For example, history-dependent adversary behaviour would require a
different definition of sub-MDPs $\online{\mdp}$, but such behaviour would not affect the size of sub-MDPs.
Hence, it would lead to similar computation times.
The MDP size and the computation times generally depend on the number of decision locations, the horizon, and the number of adversaries.
}
% Rev1, Appr2

\subsection[Safety-Relevant MDP M]{Safety-Relevant MDP $\mdp$}

In the following, we describe the safety-relevant MDP $\mdp$ underlying the  agents operating within an arena. This MDP includes non-deterministic choices of the avatar and stochastic behaviour of the adversaries.
Note that the safety-relevant MDP $\mdp$ is never explicitly created for online shielding, but is explored on-the-fly for the safety analysis of tasks.
\response{We follow the presentation by Jansen et al.~\cite{DBLP:conf/concur/0001KJSB20} for this purpose.} %Rev2, Pg6 Ln 13R

Let $G=(V,E)$ be an arena, let $\Task$ be a task function for $G$, let $B_i$ with $i\in \{1\ldots m\}$ be the behaviour functions of $m$ adversaries, and let the avatar be the zeroth agent. The safety-relevant MDP $\MdpInitR$ models the arena and agents' dynamics as follows. Each agent has a \emph{position} and a \emph{task queue} containing the activities to be performed from the last chosen task.
The agents take turns performing activities from their respective task queue.
If the task queue of an agent is empty, a new task has to be selected.
\response{Since we control the avatar, we model 
its choice of tasks as non-deterministic choice from 
all available tasks. Therefore, we analyse
the outcomes of carrying out all possible tasks. 
By Definition~\ref{def:adv_behaviour} the adversary 
behaviour is probabilistic, i.e., the adversaries choose tasks according
to a discrete probability distribution. 
}
% The avatar chooses non-deterministically, whereas the adversaries choose probabilistically. 

Hence, $\mathcal{M}$ has three types of states: (1) states where the avatar's task queue is empty and the avatar makes a \emph{non-deterministic} decision on its next task, (2) states where an adversary's task queue is empty and the adversary selects its next task \emph{probabilistically}, and (3) states where the currently active agent has a non-empty task queue and the agent processes its task queue \emph{deterministically}.

Formally, the \emph{states} $\mathcal{S}=V^{m+1} \times (E^*)^{m+1} \times \{0,\hdots, m\}$ are triples $s = (v,q,t)$ where $v$ encodes the agent positions, $q$ encodes the task queue states of all agents, and $t$ encodes whose turn it is. To enhance readability, we use $\position{s} = s[0] = v$, $\taskq{s} = s[1] = q$, and $\turn{s} = s[2] = t$ to access the elements of a state $s$. We additionally define $\avatar = 0$, thus $\position{s}[\avatar]$ and $\taskq{s}[\avatar]$ are the position and task of the avatar, whereas $\turn{s} = \avatar$ specifies that it is the turn of the avatar.
There is a unique action $\alpha_\mathrm{adv}$ representing adversary decisions, there is a unique action $\alpha_\mathrm{e}$ representing individual activities (movement along edges), and there are actions for each task available to the avatar, thus $\Act = \{\alpha_\mathrm{adv},\alpha_\mathrm{e}\} \cup \Task(G)$. 

\begin{definition}[Decision State]
Given a  safety-relevant MDP $\mdp$. 
We define the set of \emph{decision states} $\mathcal{S}_D \subseteq \mathcal{S}$ via $\mathcal{S}_D = \{s_D \in \mathcal{S}\mid \taskq{s_D}[\avatar] = \epsilon \land \turn{s_D} = \avatar\}$,
i.e., it is the turn of the avatar and its task queue is empty.
\end{definition}
This implies that if $s_D\in \mathcal{S}_D$, then $\position{s_D}[\avatar]$ is a decision location in $V_D$.
A policy for $\mathcal{M}$ needs to define actions only for states in $\mathcal{S}_D$, thereby defining the decisions for the avatar. All other task decisions in states $s$, where $\turn{s} \neq \avatar$, are performed stochastically by adversaries and cannot be controlled. 
% @COMMMENT by reviewer 1 (I think he did not understand the MDP interpretation completely ("It is not clear that handling m > 1 is more interesting than m = 1")

%
At run-time, in each turn each agent performs two steps:
\begin{compactenum}[(1)]
    \item If its task queue is empty, the agent has to select its next task and adds it to the task queue.
    \item The agent performs the next activity of its current task queue.
\end{compactenum}

\paragraph{Selecting a New Task} 
A new task has to be selected in all states $s$ with $\turn{s} = i$ and $\taskq{s}[i] = \epsilon$, i.e, it is the turn of agent $i$ and agent $i$'s task queue is empty.

If $i = \avatar$, the avatar is in a decision state  $s \in \mathcal{S}_D$, with actions $\Act(s) = \Task(\position{s}[\avatar])$. For each task $t \in \Act(s)$, there is a successor state $s'$ with $\taskq{s'} = \taskq{s}[\avatar \gets t]$, $\position{s'} = \position{s}$, $\turn{s'} = \turn{s}$, and $\pmdp(s,t,s') = 1$. Thus, there is a transition that updates the avatar's task queue 
with the edges of task $t$ with probability one. Other than that, there are no changes. 

If $i\neq \avatar$, an adversary makes a decision, thus $\Act(s) = \alpha_\mathrm{adv}$. For each $t \in \Task(\position{s}[i])$, there is a state $s'$ with $\taskq{s'} = \taskq{s}[i \gets t]$, $\position{s'} = \position{s}$, $\turn{s'} = \turn{s}$, and $\pmdp(s,\alpha_\mathrm{adv},s') = B_i(\position{s}[i])(t)$.
There is a single action with a stochastic outcome determined according to the adversary behaviour $B_i$.

\paragraph{Performing Activities} After potentially selecting a new task, the task queue of agent $i$ is non-empty. We are in a state 
$s'$, where $\taskq{s'}[i] = t = (v_i, v_i') \cdot t'$ with $\position{s'}[i] = v_i$. Agent $i$ moves along the edge $(v_i, v_i')$ deterministically and we increment the turn counter modulo $m+1$, i.e., $\Act(s') = \{\alpha_\mathrm{e}\}$ and $\pmdp(s',\alpha_\mathrm{e},s'') = 1$ with $s''[0] = \position{s'}[i \gets v_i']$, $\taskq{s''} = \taskq{s'}[i \gets t']$, and $\turn{s''} = \turn{s'} + 1 \mod m+1$.

\subsection[Sub-MDP for Next Decision]{Sub-MDP $\online{\mdp}$ for Next Decision}
\label{sec:shield_construction}

The idea of online shielding is to compute the safety value of actions in the decision states on 
the fly and block actions that are too risky. 
For infinite horizon properties, the probability to violate safety, in the long run, is often one and errors stemming from modelling uncertainties may sum up over time~\cite{DBLP:conf/concur/0001KJSB20}.
Therefore, we consider safety relative to a \emph{finite horizon} such that the action values 
(and consequently, a policy for the avatar) carry guarantees for the next several steps. 
Explicitly constructing an MDP $\mathcal{M}$ as outlined above yields a very large number of decision states that may be infeasible to check. 
The finite horizon assumption allows us to prune the safety-relevant MDP
and construct small sub-MDPs $\online{\mathcal{M}}$ capturing the immediate future of individual decision states.

More concretely, we consider runtime situations of being in a state $s_t$, the state visited immediately after the avatar decided to perform a task $t$. In such situations, we can use the time required to perform $t$ for shield computations for the next decision. 
We create a sub-MDP $\online{\mathcal{M}}$ by determining all states reachable within a finite horizon and use
$\online{\mathcal{M}}$ to check the safety probability of each action (task) available in the next decision
and block unsafe actions. 

\paragraph[Construction of sub-MDP]{Construction of $\online{\mdp}$}

Online shielding relies on the insight that after deciding on a task $t$, the time required to complete $t$ can be used to compute a shield for the next decision. Thus, we start the construction of the sub-MDP $\online{\mdp}$ for the next decision location $v_D'$ from the state $s_t$ that immediately follows a decision state $s_D$, where the avatar has chosen a task $t \in \Act(s_D)$. The MDP $\online{\mdp}$ is computed with respect to a finite horizon $h$ for $v_D'$. 

By construction, the task is of the form $t = v_D \cdots v_D'$, where $v_D$ is the avatar's current location and $v_D'$ is the next decision location. 
While the avatar performs $t$ to reach $v_D'$, the adversaries perform arbitrary tasks and traverse $|t|$ edges, i.e., until $v_D'$ is reached only adversaries make decisions.
This leads to a set of possible next decision states. We call these states the \emph{first decision states} $S_\mathrm{FD} \subseteq \mathcal{S_D}$. 
After reaching $v_D'$, both avatar and adversaries decide on arbitrary tasks and all agents traverse $h$ edges. 
This behaviour defines the structure of $\online{\mdp}$.

Given a safety-relevant MDP $\MdpInitR$, a decision state $s_D$ and its successor $s_t$ with $\taskq{s_t}[\avatar] = t$,
and a finite horizon $h \in \mathbb{N}$ representing a number of turns taken by all agents following the next decision. 
These turns and the (stochastic) agent behaviour leading to the next decision are modelled by the sub-MDP $\shielded{\mdp}$.
$\shielded{\mdp}=(\shielded{\mathcal{S}},\shielded{{s_0}},\shielded{\Act},\shielded{\pmdp} )$ is formally constructed as follows. 
The actions are the same as for $\mdp$, i.e., $\shielded{\Act}=\Act$. The initial state is given by $\shielded{{s_0}}=(s_t,0)$.
The states of $\shielded{\mdp}$ are a subset of $\mdp$'s states augmented with the distance from $\shielded{{s_0}}$, i.e., $\online{\mathcal{S}} \subseteq \mathcal{S} \times \mathbb{N}_0$. The distance is measured in terms of the number of turns taken by all agents.

We define transitions and states inductively by: 
\begin{compactenum}[(1)]
\item{\sl Decision Actions.} If $(s,d) \in \online{\mathcal{S}}$, $d < |t| + h$, and there is an $s'\in \mathcal{S}$ such that $\pmdp(s,\act,s') > 0$ and $\act \in \{\act_\mathrm{adv}\} \cup \Task(G)$ then  $(s',d) \in \online{\mathcal{S}}$ and $\shielded{\pmdp}((s,d),\act,(s',d)) = \pmdp(s,\act,s')$. 
\item{\sl Movement Actions.} If $(s,d) \in \online{\mathcal{S}}$, $d < |t| + h$, and there is an $s' \in \mathcal{S}$ such that $\pmdp(s,\act_\mathrm{e},s') > 0$, then  $(s',d') \in \online{\mathcal{S}}$ and $\shielded{\pmdp}((s,d),\act_\mathrm{e},(s',d'))  = \pmdp(s,\act_\mathrm{e},s')$, where $d' = d + 1$ if $\turn{s} = m$ and $d' = d$ otherwise. 
\end{compactenum} 
Movements of the last of $m+1$ agents increase the distance from the initial state. Combined with the fact that every movement action increases the agent index and every decision changes a task queue, we can infer that the structure of $\online{\mdp}$ is a directed acyclic graph. This enables an efficient probabilistic analysis.

By construction, it holds that for every state $(s,d) \in \online{\mathcal{S}}$ with $d < |t|$, $s$ is not a decision state of $\mdp$.
The set of first decision states $S_\mathrm{FD}$ consists of all states $s_\mathrm{FD} = (s, |t|)$ such that $s_\mathrm{FD} \in \online{\mathcal{S}}$ with $\taskq{s}[\avatar] = \epsilon$ and $\turn{s} = \avatar$, i.e., all first decision states reachable from the initial state of $\online{\mdp}$. 
We use $\Task(S_\mathrm{FD}) = \{t \mid s \in S_\mathrm{FD}, t \in \Act(s)\}$ to denote the tasks available in these states. $\shielded{\mdp}$ does not define actions and transitions from states $(s,|t| + h) \in \online{\mathcal{S}}$, as their successor states are beyond the considered horizon $h$. 
%We have $\Act((s,|t| + h)) \neq \emptyset$ for all states at distance $|t| + h$ from the initial state.
%BK: ask Martin
We have $\Act((s, d)) \neq \emptyset$ for all states at distance $d < |t| + h$ from the initial state.

\subsection{Shield Construction}

The probability of reaching a set of unsafe states $T \in \mathcal{S}$ from any state in the safety-relevant MDP should be low.
In the finite horizon setting, we are interested in bounded reachability from decision states $s_D \in \mathcal{S}_D$ within the finite horizon $h$.
We concretely evaluate reachability on sub-MDPs $\online{\mdp}$ and use $\online{T} = \{(s,d) \in \online{S} \mid s \in T\}$ to denote the unsafe states that may be reached within the horizon covered by $\online{\mdp}$.
The property $\varphi=\finally \online{T}$ encodes the \textbf{violation} of the safety constraint, i.e., eventually reaching $\online{T}$ within $\online{\mdp}$.
The shield needs to limit the probability to \emph{satisfy} $\varphi$.

Given a sub-MDP $\shielded{\mdp}$ and a set of first decision states $S_\mathrm{FD}$. For each task $t \in Task(S_\mathrm{FD})$,
we evaluate $t$ with respect to the minimal probability to satisfy $\varphi$ from the initial state $\shielded{{s_0}}$ when executing $t$ 
\response{that is given by} $\mcresmin{\varphi}{\online{\mdp}}(\shielded{{s_0}})$. This is formalised with the notion of task-valuations below.

\begin{definition}[Task-valuation]\label{def:actionvalues}
	A \emph{task-valuation} for a task $t$ in a sub-MDP $\online{\mdp}$ with initial state $\shielded{{s_0}}$ and first decision states $S_\mathrm{FD}$ is given by
	\begin{align*}
 \val{\online{\mdp}}&\colon \Task(S_\mathrm{FD}) \rightarrow [0,1] \text{, } \\
 &\text{with } 
 \val{\online{\mdp}}(t) = \mcresmin{\varphi}{\online{\mdp}}(\shielded{{s_0}})\text{,} \\
 &\text{and } \Act(s_\mathrm{FD}) = \{t\} \text{ for each } s_\mathrm{FD} \in S_\mathrm{FD}\text{.}
\end{align*}
The \emph{optimal task-value} for $\online{\mdp}$ is $\optval{\online{\mdp}} =  \min_{t' \in \Task(S_\mathrm{FD})} \val{\online{\mdp}}(t')$.
\end{definition}

A task-valuation is the minimal probability to reach an unsafe state in $T$ from each immediately reachable decision state $s_\mathrm{FD} \in S_\mathrm{FD}$ weighted by the probability to reach $s_\mathrm{FD}$. 
When the avatar chooses an optimal task $t$ (with $\val{\online{\mdp}}(t) = \optval{\online{\mdp}}$) as next task in a state $s_\mathrm{FD}$, $\optval{\online{\mdp}}$ can be achieved if all subsequent decisions are optimal as well.

We now define a shield for the decision states $S_\mathrm{FD}$ in a sub-MDP $\online{\mdp}$ using the task-valuations.
Specifically, a \emph{shield} for a threshold $\delta\in[0,1]$ determines a set of  tasks available in $S_\mathrm{FD}$  that are $\delta$-optimal for the specification $\varphi$. All other tasks are ``shielded'' or ``blocked''.

\begin{definition}[Shield]\label{def:shield}
For task-valuation $\val{\online{\mdp}}$ and a threshold $\delta \in [0,1]$, a \emph{shield for $S_\mathrm{FD}$ in \online{\mdp}} is given by
\begin{align*}
	\shield{\delta}{\online{\mdp}} &\in 2^{\Task(S_\mathrm{FD})} \text{ with } \\
	\shield{\delta}{\online{\mdp}} &= \{ t \in \Task(S_\mathrm{FD}) \mid  \\ 
	& \qquad \delta \cdot \val{\online{\mdp}}(t) \leq \optval{\online{\mdp}} \}\text{.}
\end{align*}
\end{definition}

Intuitively, $\delta$ enforces a constraint on tasks that are acceptable w.r.t. the optimal probability.
%\ie, which actions are $\delta$-optimal.
The shield is \emph{adaptive} with respect to $\delta$, as a high value for $\delta$ yields a stricter shield, a smaller value a more permissive shield.
% The shield is stored using a lookup-table, and the value for $\delta$ can then be changed on-the-fly.
In particularly critical situations, the shield can enforce the decision maker to resort to (only) the optimal actions w.r.t. the safety objective. 
This can be achieved by temporarily setting $\delta = 1$. Online shielding creates shields on-the-fly by constructing sub-MDPs $\online{\mdp}$ and computing task-valuations for all available tasks.

Through online shielding, we transform the safety-relevant MDP $\mdp$ into a \emph{shielded MDP} with which the avatar interacts (which is never explicitly created) that is obtained from the composition of all sub-MDP$\shielded{\mdp}$.
Due to the assumption on the task functions that requires a non-empty set of available tasks in all decision locations
and due to the fact that every decision for shielding is defined w.r.t. an optimal task, the shielded MDP
is deadlock-free. \response{The deadlock-freedom follows from
using a relative threshold ensuring that the safest action never gets blocked,
thus at least one action is always available.
Using task valuations as a basis, our notion of online shielding guarantees optimality with respect to safety.} 
%Hence, our notion of online shielding guarantees optimality w.r.t. safety and deadlock-freedom. 

By using the minimal probability as task valuation $\val{\online{\mdp}}(t)$, we assume that the avatar performs optimally with respect to safety in upcoming decisions. \response{This means that to compute task valuations we assume that the agent always chooses the safest action that is available. Alternative definitions that are less
optimistic would be possible as well. Defining 
task valuations as the maximal probability to violate
safety would yield stricter guarantees. The corresponding shield would block each action $a$ if there
exists any future behaviour following $a$ that violates
safety with a too large probability.}

%To implement stricter shielding,
\response{Instead of a relative threshold, }
we may use a fixed absolute threshold $\lambda \in [0,1]$ such that only tasks $t$ with $\val{\online{\mdp}}(t)\leq\lambda$ are allowed.
Since this may induce deadlocks in case there are no sufficiently safe actions, we fall back to shielding with $\delta=1$, which defines the threshold as the valuation of the safest action. That is, if there are no $\lambda$-safe actions, we use the safest action available. \response{Hence, we either enforce a limit on unsafe behaviour or use the safest 
option available, which avoids deadlocks. }
% @COMMENT of reviewer 2 

\subsection{Optimisation -- Updating Shields after Adversary Decisions}
\label{sec:shield_opt}

After the avatar decides on a task, we use the time to complete the task to compute shields based on task-valuations (see \cref{def:actionvalues} and \cref{def:shield}). Such shield computations are inherently affected by uncertainties stemming from stochastic adversary behaviour. These uncertainties consequently decrease whenever we observe a concrete decision from an adversary that we considered stochastic in the initial shield computation. 

An optimisation of the online shielding approach is to compute a new shield after any decision of an adversary,
if there is enough remaining time until the next decision location.
Suppose that after visiting a decision state, we computed a shield based on $\shielded{\mdp}$.
While moving to the next decision state, an adversary decides on a new task and we observe the concrete state $s$.
We can now construct a new sub-MDP $\online{\mdp'}$ using $\shielded{{s_0'}}=s$ as initial state, thereby resolving a stochastic decision between the original initial state $\shielded{{s_0}}$ and $\shielded{{s_0'}}$. 
Using $\online{\mdp'}$, we compute a new shield for the next decision location.

The facts that the probabilistic transition function of $\online{\mdp}$ does not change during updates and that we consider safety properties enable a very efficient implementation of updates. For instance, if value iteration is used to compute task-valuations, we can simply change the initial state and reuse computations from the initial shield computation. 
Note that if a task is completely safe, i.e., tasks with a valuation of zero, the value of this task will not change under a re-computation, since the task is safe under any sequence of adversary decisions.

\section{Implementation and Experiments}\label{sec:experiments}
We consider several experiments of shielded reinforcement learning for a 2-player version of the classic game \snake.
We picked this game because it requires fast decision making during runtime, and
provides an intuitive and fun setting to show the potential of shielding such that it can potentially be used for teaching formal methods. 
Furthermore the game is interesting for shielding, since the agent has to experience risky situations in order to win the game. This allows us to evaluate the ability of the shield in protecting the agent from entering unsafe states, to study the influence of the shield on the learning performance, and to evaluate the safety of the final learned policy. 

In this section, we start by giving details on the computation of the shield. In this context, we examine the runtime of shield computations. Afterwards, we discuss several experiments of shielding learning agents in tabular Q-learning and deep Q-learning.

The source code can be found at \url{http://onlineshielding.at} along with videos, evaluation data, and a Docker image that enables easy experimentation.

\subsection{2-Player \snake}
\begin{figure}[t]
\centering
\includegraphics[width=.45\textwidth]{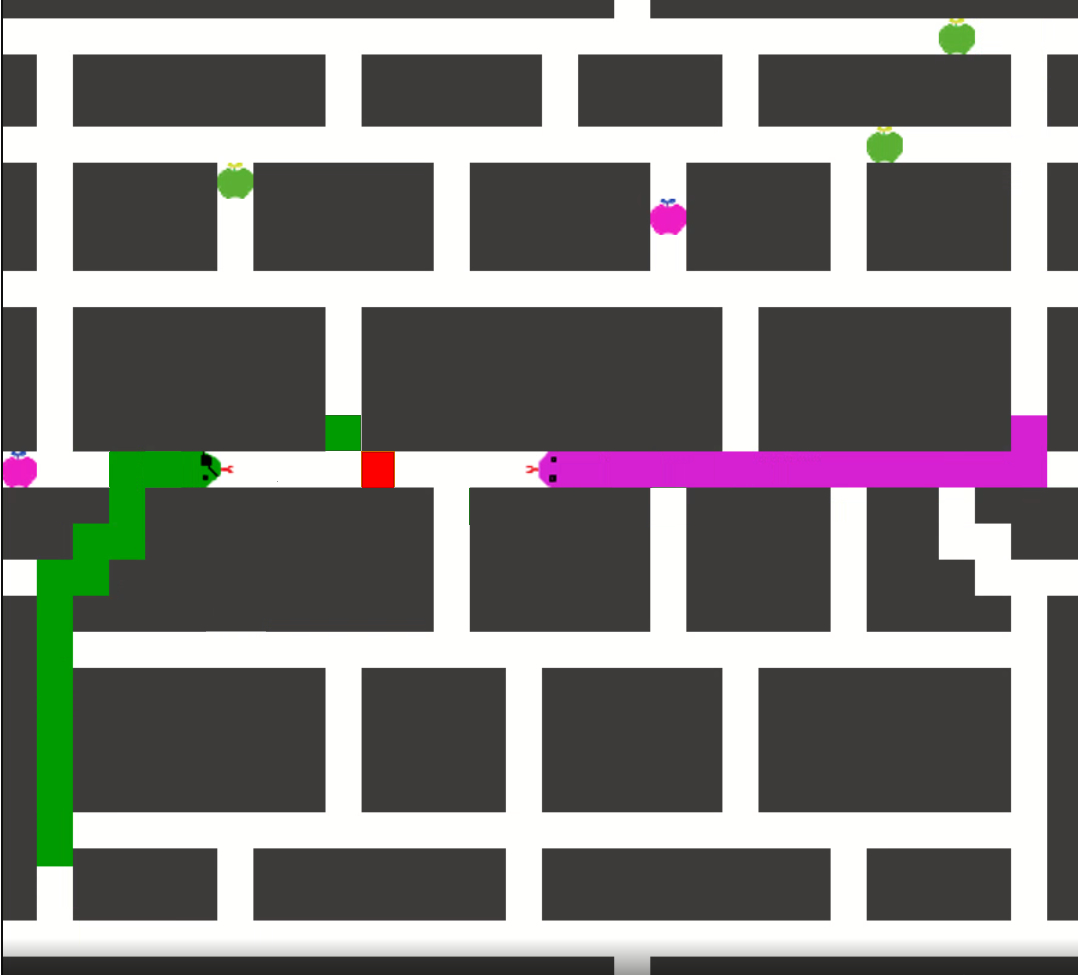}
\caption{A screenshot of the \snake game using Map 1 with colour-coded shield display.}
\label{fig:sceenshot1}
\end{figure}

In the game of 2-player \snake, each player controls a snake of a different colour. 
A player wins, if it is either able to eat all randomly positioned apples of their own colour before the adversary snake collects all apples in its colour, or if it is able to cut off the other player.
In the case that the heads of both snakes collide, the game ends in a tie.

We provide an open-source implementation of the 2-player \snake game.
The game can be played on 6 different maps, one of which is shown in 
Figure~\ref{fig:sceenshot1}. The game settings allow varying the lengths of the snakes and their speed, allowing the user to set different levels of difficulty. For both snakes, shielding can be activated and deactivated.
The game can be played in the following player modes: (1) double player (two human players compete), (2) player against agent (human player plays against a trained learning agent), or (3) agent against agent. The third mode is used for reinforcement learning. One snake is controlled by the avatar (the RL-agent), and the adversary snake is controlled by a trained agent.

\textbf{Implementation.} The game's interface and logic was implemented using the pygame\footnote{\url{https://www.pygame.org/}, \DTMdisplaydate{2020}{11}{27}{-1}} library. 

\subsection{Shield Computation}
\begin{figure}[t]
\centering
\includegraphics[width=0.47\textwidth]{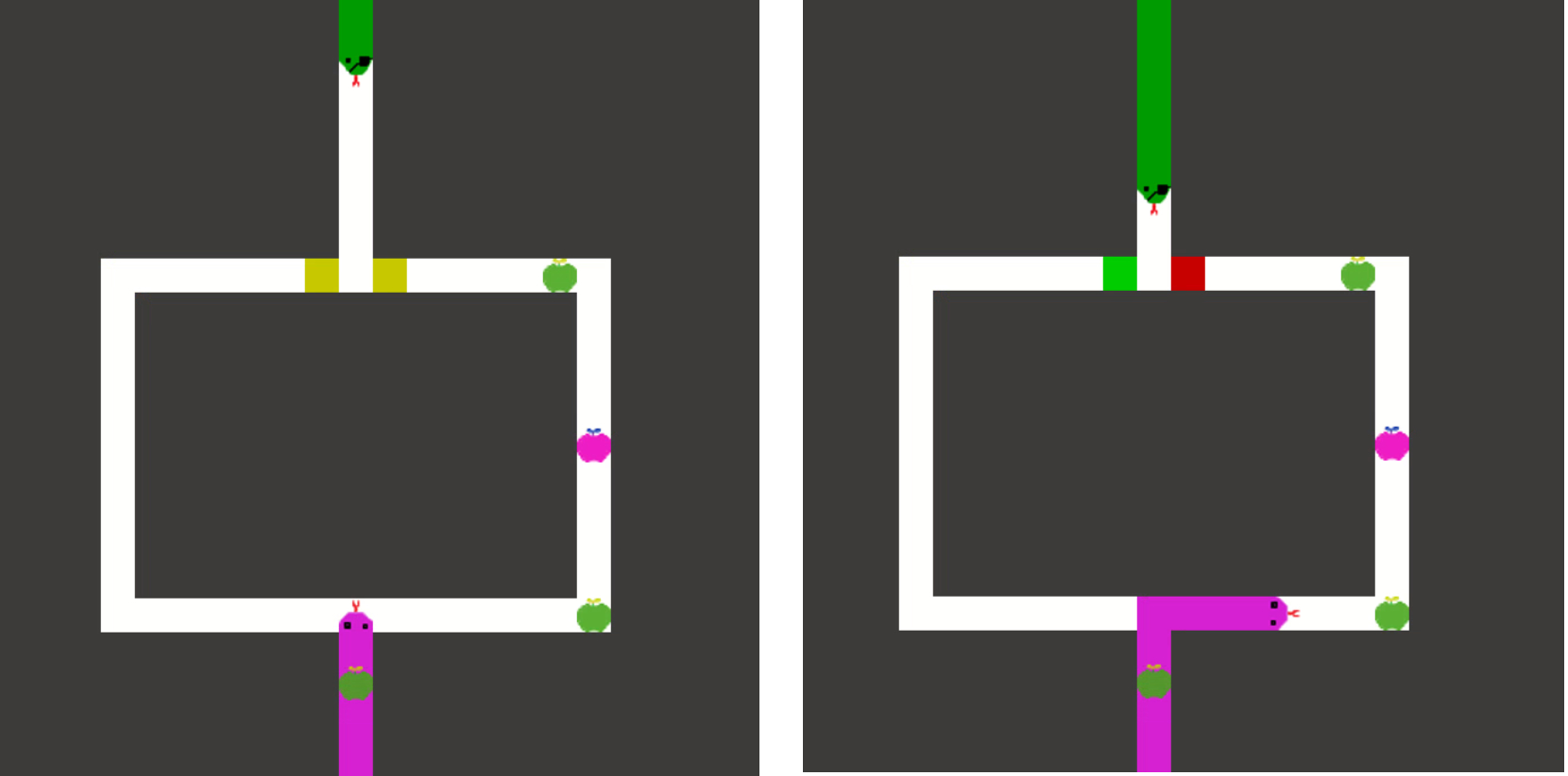}
\caption{Screenshots from the \snake game using a simple map to demonstrate recalculation.}
\label{fig:sceenshot_shield_and_recalc}
\end{figure}

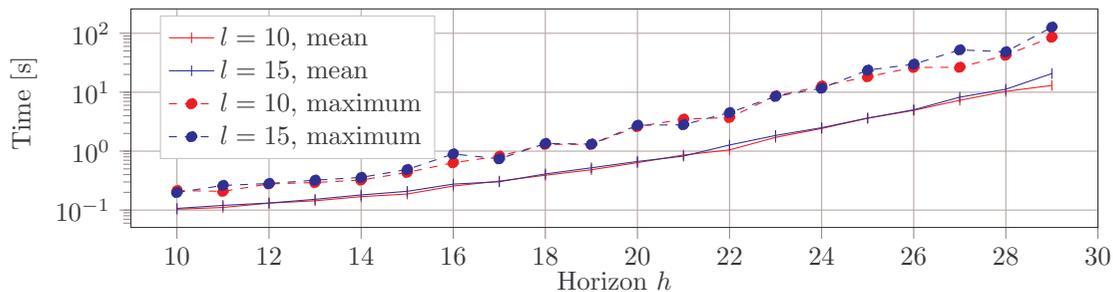
\begin{figure*}[t]
    \centering
    \begin{tikzpicture}
\begin{axis}[legend cell align={left},
legend entries={{$l=10$, mean},{$l=15$, mean},{$l=10$, maximum},{$l=15$, maximum}},
legend style={at={(0.03,0.97)}, anchor=north west, draw=white!80.0!black},
tick align=outside,
tick pos=left,
x grid style={lightgray!92.02614379084967!black},
xlabel={Horizon $h$},
xlabel style = {below=2mm},
ylabel style = {above=2mm},
title style = {above=3mm},
xmajorgrids,
xmin=9, xmax=30,
y grid style={lightgray!92.02614379084967!black},
ylabel={Time [s]},
ymajorgrids,
width=0.9\textwidth,
height=0.28\textwidth,
x label style={at={(axis description cs:0.5,-0.09)},anchor=north},
ymode=log]

\addplot[color=red, style = solid, mark = |] coordinates { %10-mean
(10,0.10339624687447213)
(11,0.11077590154571226)
(12,0.1310319801006699)
(13,0.14368726900283946)
(14,0.16911989896558224)
(15,0.18721914495923556)
(16,0.2554155680484837)
(17,0.3106202617267263)
(18,0.38990108019876063)
(19,0.48311986114567845)
(20,0.6370980332864565)
(21,0.8593174993235152)
(22,1.0488577481624088)
(23,1.713255172832287)
(24,2.4239968288896487)
(25,3.622418735098327)
(26,4.922773461265024)
(27,7.30723524538509)
(28,10.343160685923358)
(29,13.030479521512461)
% (30,20.687567864372685)
};
\addplot[color=blue, style = solid, mark = |] coordinates { %15-mean
(10,0.10665499687544071)
(11,0.12013743004150455)
(12,0.13175734899123198)
(13,0.1521771593161975)
(14,0.1807794839469716)
(15,0.20805640439968556)
(16,0.27587092314293843)
(17,0.30226686300506117)
(18,0.4104779734439217)
(19,0.5215225122190895)
(20,0.6681664935959271)
(21,0.8234064633597155)
(22,1.271768888492079)
(23,1.8452586279041134)
(24,2.4979558239213655)
(25,3.6438199609945876)
(26,5.036405216602143)
(27,8.23255051202228)
(28,11.235037990765704)
(29,20.738799952500557)
};
\addplot[color=red, style = dashed, mark = *] coordinates { %10-wc
(10,0.21436117397388443)
(11,0.2079704569769092)
(12,0.2780317129800096)
(13,0.2936996130156331)
(14,0.3228294490254484)
(15,0.4350784639827907)
(16,0.6355329009820707)
(17,0.8210048309992999)
(18,1.3053794450243004)
(19,1.3036258540232666)
(20,2.610773002030328)
(21,3.4872464640066028)
(22,3.7140747309895232)
(23,8.657626482017804)
(24,12.714245119015686)
(25,18.282771444995888)
(26,26.35125045699533)
(27,26.49520757497521)
(28,42.40304919099435)
(29,85.78463311999803)
% (30,71.22069477598416)
};
\addplot[color=blue, style = dashed, mark = *] coordinates { %15-wc
(10,0.19867584604071453)
(11,0.2620728629990481)
(12,0.282208206015639)
(13,0.32163305598078296)
(14,0.35842895798850805)
(15,0.48769047000678256)
(16,0.8963309100363404)
(17,0.7393665809649974)
(18,1.349632297991775)
(19,1.3153385600307956)
(20,2.748471908038482)
(21,2.806087204022333)
(22,4.5055409540073015)
(23,8.473888397973496)
(24,11.683173689001705)
(25,23.68352826498449)
(26,29.85266632999992)
(27,52.281756129988935)
(28,48.5033698239713)
(29,127.10364941897569)
};

\end{axis}
\end{tikzpicture}
    \caption{Shield computation time for varying horizon values and snake lengths.}
    \label{fig:eval_time}
\end{figure*}
The task of the shield is to protect its snake from collisions with the adversary snake and with its own body. 

\response{In the safety-relevant MDP $\mathcal{M}$,
the avatar snake can make a decision at every crossing,
thus the crossings define the decision locations.
The states in $\mathcal{M}$ store the positions of the bodies of both snakes.
Therefore, we store for each snake the location of the head, the tail, and all crossings that are covered by the body of the snake. The locations of the corridors covered by the bodies are then implicitly defined.
The safety specification $\varphi$ defines that the heads of the snake should not collide and that the head of the avatar snake should not crash into the body of the adversary snake
or its own body. }
\response{This results in a safety specification $\varphi = \mathbf{G}(\neg \texttt{Collision\_Heads}(\texttt{head\_s1}, \texttt{head\_s2}) $ $\wedge 
\neg \texttt{Collision\_Bodies}(\texttt{head\_s1}, \{\texttt{body\_s1}\},$ $\texttt{tail\_s1},$ $\{\texttt{body\_s2}\}$, $\texttt{tail\_s2}))$, where the predicates $\texttt{Collision\_Bodies}$ and $\texttt{Collision\_Heads}$ compare the locations of the snakes and check for collisions.}
\response{Each time the avatar snake enters a corridor,
the shielding approach creates sub-MDPs $\online{\mdp}$ for the next crossing and possible future positions for the adversary snake. 
Given such a sub-MDP $\online{\mdp}$ and the safety property $\varphi$, we compute the minimal probability to violate $\varphi$ within the next $h$ steps. When the avatar snake arrives at the next crossing, 
\emph{the shield allows only the corridors with the highest safety value}}.
The game, as shown in Figure~\ref{fig:sceenshot1}, indicates the risk of taking a corridor from low to high by the colours green, yellow, orange, red. 

We also implemented the optimisation to recalculate the shield after a decision of the adversary snake.
Figure~\ref{fig:sceenshot_shield_and_recalc} contains two screenshots of the game on a simple map to demonstrate 
the effect of a shield update. In the left figure, the available tasks of the green snake 
are picking the corridor to the left or the corridor to the right.
Both choices induce a risk of a collision with the purple snake. After the decision of 
the purple snake to take the corridor to its right-hand-side, the shield is updated and the 
safety values of the corridors change.

\textbf{Experimental Set-up.} The shield computation uses the probabilistic model checker \storm~\cite{DJKV17} and its Python interface to compute the safety of actions. We use the \prism~\cite{DBLP:conf/cav/KwiatkowskaNP11} language to represent MDPs and domain-specific optimisations to efficiently encode agents and tasks, that is, snakes and their movements.
The experiments with tabular Q-learning have been performed on a computer with an Intel\textsuperscript{\textregistered{}}Core\texttrademark{} i7-4700MQ\footnote{Intel and the Intel logo are trademarks of Intel Corporation or its subsidiaries.} CPU with 2.4 GHz, 8 cores and 16 GB RAM. 
All experiments with deep Q-learning have been performed on a computer with an Intel\textsuperscript{\textregistered{}}Core\texttrademark{} i5-6600 CPU with 3.3 GHz, 4 cores, an NVIDIA\textsuperscript{\textregistered{}} GeForce\textsuperscript{\textregistered{}} GTX 1070 and 16 GB RAM.

\textbf{Runtime Measurements.}
% \textbf{Results.} 
We report the time required to compute and analyze the safety of actions (i.e., to compute the shield) relative to the computation horizon.
The experiments on computation time indicate how many steps shielding can look ahead within some given time.

When playing the game on Map 1 illustrated in Figure~\ref{fig:sceenshot1}, we measured the time to compute shields, i.e., the time to construct sub-MDPs $\shielded{\mdp}$ and to compute the safety values.
We measured the time of $200$ such shield computations and report the maximum computation times and the mean computation times.
Figure~\ref{fig:eval_time} presents the results for two different snake lengths $l \in \{10, 15\}$ and
different computation horizons $h \in \{10, 11, \dots, 29\}$. 
The x-axis displays the computation horizon $h$ and the y-axis displays the computation time in seconds in logarithmic scale.

We can observe that up to a horizon $h$ of $17$, all computations take less than one second, even in the worst case. Assuming that every task takes at least one second, we can plan ahead by taking into account safety hazards within the next $17$ steps. A computation horizon of $20$ still requires less than one second on average and about $3$ seconds in the worst case. Horizons in this range are often sufficient, as we demonstrate in the next experiment by using $h=15$. 

We compare our timing results with a similar case study presented by Jansen et al.~\cite{DBLP:conf/concur/0001KJSB20}.
In a similar multi-agent setting on a comparably large map, 
the decisions of the avatar were shielded using an offline shield
with a finite horizon of $10$. The computation time to compute the offline shield
was about \emph{6 hours} on a standard notebook.
Note, that although the setting has four adversaries, the offline computation was performed for one adversary and the results were combined for several adversaries online.

Furthermore, Figure~\ref{fig:eval_time} shows that the snake length affects the computation time only slightly. This observation supports our claim that online shielding scales well to large arenas, i.e., scenarios where the safety-relevant MDP $\mdp$ is large. Note that the number of game configurations grows exponentially with the snake length (assuming a sufficiently large map), as the snake's tail may bend in different directions at each crossing. 

The experiments further show that the computation time grows exponentially with the horizon. 
Horizons close to $30$ may be advantageous in especially safety-critical settings, such as factories with industrial robots acting as agents. 
Since individual tasks in a factory may take minutes, online shielding would be feasible, as worst-case computation times are in the range of minutes. However, offline shielding would be infeasible due to the average computation time of more than $10$ seconds that would be required for all decision states, \response{of which there are thousands. As a result, computing an offline shield would require
days and require a large shielding database.}

\response{
\noindent\textbf{Complexity Analysis.}
To analyse the complexity of the constructed sub-MDPs, let $h$ be the 
horizon, let $n_a \leq h$ ($n_e \leq h$) be the number of decision states 
reachable within $h$ steps by the avatar (adversary), and let $l$ be the length
of the snakes. In the worst case, the avatar snake can  bend at $n_a$ points. Therefore,
the number of reachable states for the avatar snake within a horizon of $h$
is at most $h \cdot l^{n_a}$. The same holds for the
adversary snake, resulting in a state space in $O(h\cdot(l^{n_a} \cdot l^{n_e}))$. Note that adding further adversary snakes adds additional factors
$l^n$.
}

\label{sec:rewards_shielding_experiments}
\begin{figure*}[t]
    \centering
\begin{tikzpicture}
\begin{axis}[legend cell align={left},
legend entries={{Shielded RL},{Unshielded RL}},
legend style={at={(1.32,0.31)}, anchor= south east, draw=white!80.0!black},
tick align=outside,
tick pos=left,
x grid style={lightgray!92.02614379084967!black},
xlabel={Learning Episode},
xlabel style = {below=2mm},
ylabel style = {above=2mm},
title style = {above=3mm},
xmajorgrids,
y grid style={lightgray!92.02614379084967!black},
ylabel={Reward},
ymajorgrids,
width=0.79\textwidth,
height=0.30\textwidth,
x label style={at={(axis description cs:0.5,-0.09)},anchor=north},
]

\addplot [color=red, style = solid, mark = x] coordinates { %experiments with shield
(0,0)
(50,62.0)
(100,66.0)
(150,71.8)
(200,75.2)
(250,72.8)
(300,65.0)
(350,77.0)
(400,68.2)
(450,60.4)
(500,73.6)
(550,61.8)
(600,43.4)
(650,70.0)
(700,71.0)
(750,72.0)
(800,72.6)
};

\addplot [color=blue, style = solid, mark = *] coordinates { %experiments without shield
(0,0)
(50,13.0)
(100,6.4)
(150,-1.8)
(200,13.8)
(250,7.4)
(300,9.0)
(350,20.8)
(400,24.2)
(450,25.2)
(500,46.6)
(550,48.0)
(600,8.0)
(650,44.2)
(700,35.4)
(750,16.4)
(800,40.4)
};
\end{axis}
\end{tikzpicture}
    \caption{Results Agent 1: Reward gained in training.}
    \label{fig:reward_plots}
\end{figure*}
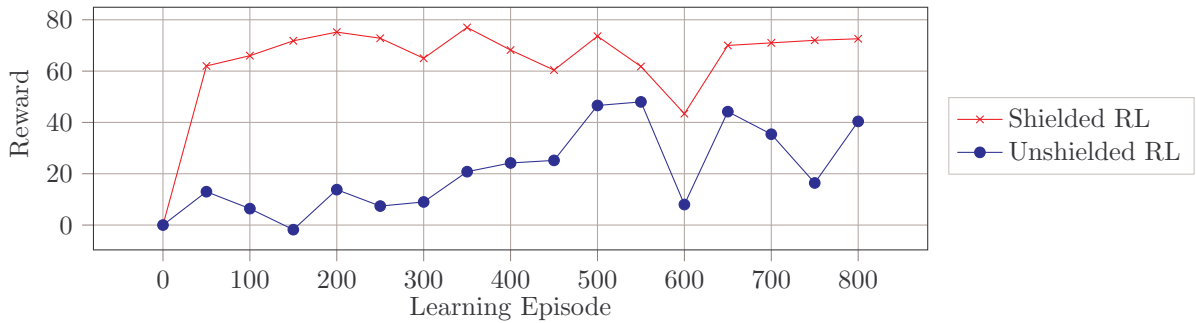
\subsection{Shielding for Tabular Q-learning}

Next, we study the effects of shielding on a simple approximate Q-learning agent~\cite{sutton1998reinforcement}, called Agent 1, that we train on Map 1 (Figure~\ref{fig:sceenshot1}).

\textbf{Learning parameters.} 
The feature vector of the approximate Q-learning agent denotes the distance to the next apple. The Q-learning uses the learning rate $\alpha = 0.1$ and the
discount factor $\gamma = 0.5$ for the Q-update and an $\epsilon$-greedy exploration policy with $\epsilon$ = 0.6.
The reward function of the RL agent is positively affected (+10) by collecting an apple in its colour and by wins of the avatar (+50), i.e., if it collects all apples in its own colour before the adversary snake collects all apples in the colour of the adversary. \response{The agent receives a reward from -100 for losing the game.}

\textbf{Results.}
To demonstrate the effects of shielding, we report the performance of shielded RL compared to unshielded RL during training measured in terms of gained reward.  

Figure~\ref{fig:reward_plots} shows plots of the reward gained during learning in the shielded and the unshielded case. The online shield uses a horizon of $h=15$. The y-axis displays the reward and the x-axis displays the learning episodes, where one episode corresponds to one play of the \snake game. The reward has been averaged over 50 episodes for each data point. 
The plot demonstrates that shielding improves the gained reward significantly. By blocking unsafe actions, 
the avatar did not encounter a single loss due to a collision. 
For this reason, we see a consistently high reward right from the start of the learning phase. 
To evaluate the performance of the learned policies, we executed 100 games with the policies obtained from unshielded RL and shielded RL, keeping the shield in place. The shielded RL agent won $96$ of all plays, whereas unshielded RL won only $54$ plays.

\begin{figure*}[t]
        \centering
    
    \begin{tikzpicture}
    \node(map){
         \includegraphics[width=0.24\textwidth]{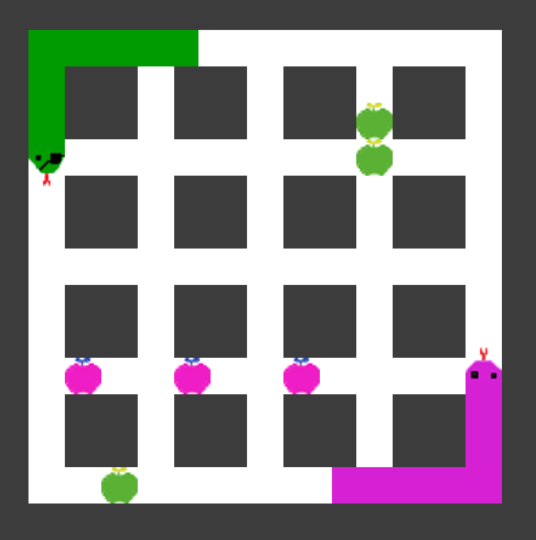}};
         \node[right = 1.2cm of map]{\includegraphics[width=0.55\textwidth]{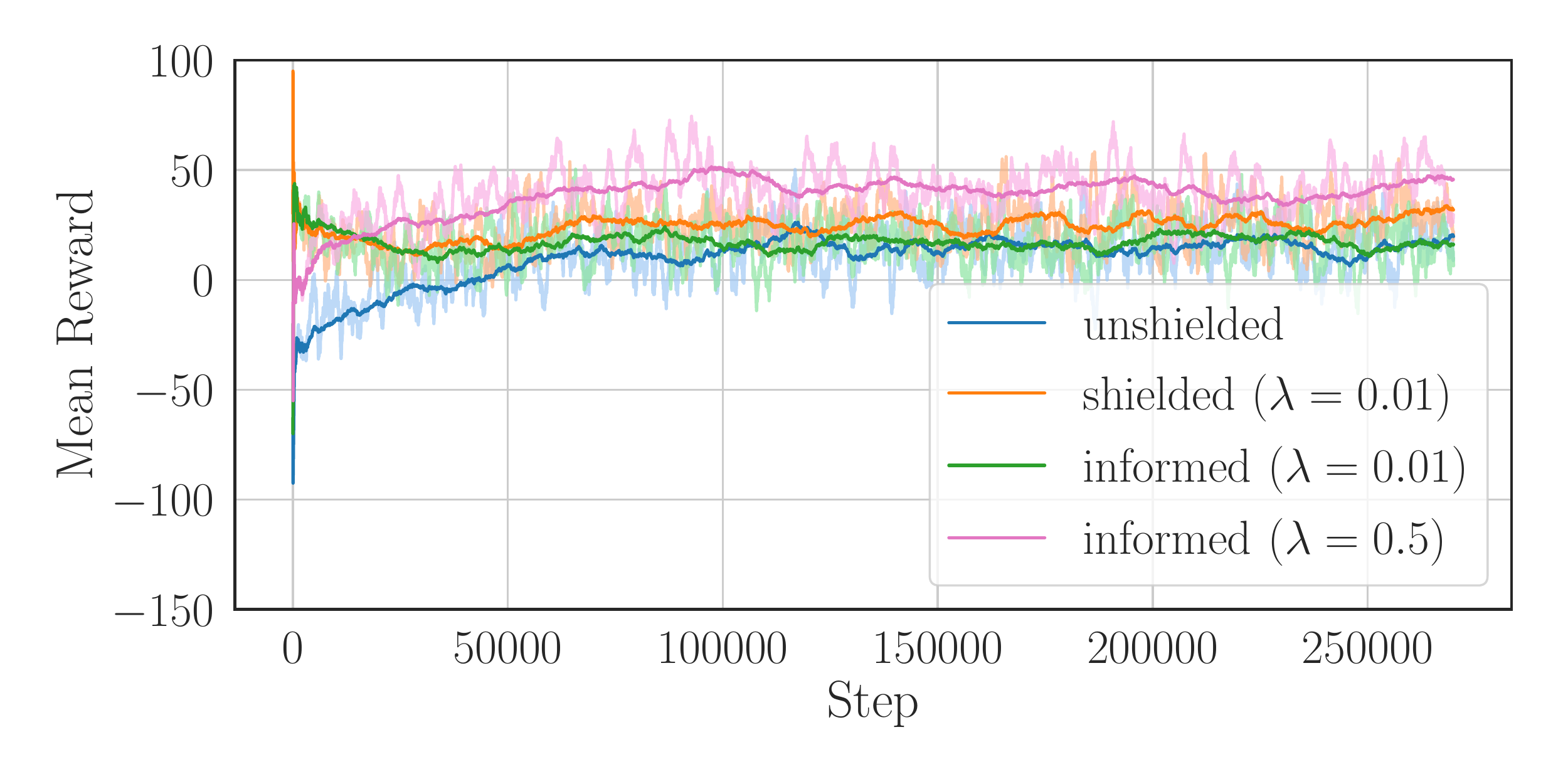}};
    \end{tikzpicture}
    
    \caption{ Results for Agent 2. 
    Left:  A screenshot of the game using Map 2;
    Right: Reward gained in training.}
    \label{fig:training_reward_plot_agent2}
\end{figure*}

\iffalse
\begin{figure*}[t]
    \centering
\scalebox{0.93}{
\begin{tabular}{|c|cc|cc|cc|cc|}
\hline
 & \multicolumn{2}{c|}{unshielded} & \multicolumn{2}{c|}{shielded ($\lambda=0.01$)} & \multicolumn{2}{c|}{informed ($\lambda=0.01$)} & \multicolumn{2}{c|}{informed ($\lambda=0.5$)} \\
 \hline
                        & \multicolumn{1}{c|}{\AVGREW} & wins & \multicolumn{1}{c|}{\AVGREW} & wins & \multicolumn{1}{c|}{\AVGREW} & wins & \multicolumn{1}{c|}{\AVGREW} & wins \\ \hline
executed unshielded    & \multicolumn{1}{c|}{26.78}   &   454   & \multicolumn{1}{c|}{-22.00}   &  687   & \multicolumn{1}{c|}{3.480}   &   847   & \multicolumn{1}{c|}{-14.44} &  713 \\ \hline
executed shielded & \multicolumn{1}{c|}{50.41}   &   809   & \multicolumn{1}{c|}{52.35}   &  835    & \multicolumn{1}{c|}{27.56}   &  803    & \multicolumn{1}{c|}{54.55} &  797 \\ \hline
\end{tabular}}
    \caption{Execution Results for Agent 2.}
    \label{fig:execution_results_agent2}
\end{figure*}
\fi

\begin{figure*}[t]
    \centering
    \includegraphics[width=1\textwidth]{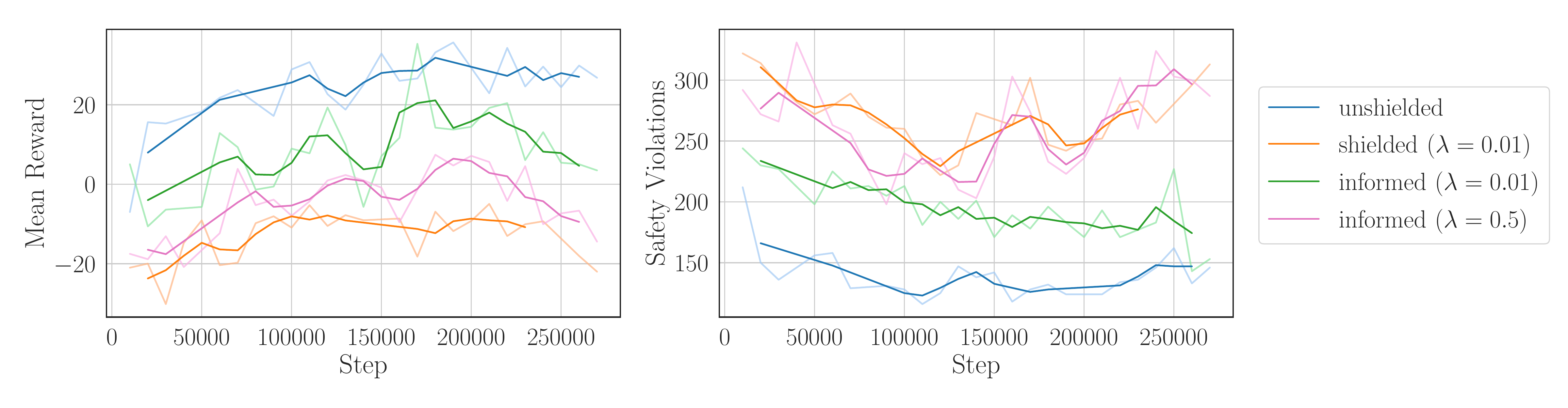}
    \caption{Intermediate execution results for Agent 2.
    Left: Reward.
    Right: Safety violations. }
    \label{fig:execution_reward_plot_agent2}
\end{figure*}
\subsection{Shielding for Deep Q-Learning}

In the next experiments, we extend the simple tabular Q-learning
setting into deep Q-Learning. 
The goal is to study the effects of shielding on the learning performance and the safety of the final policy,
if the shield actively alters the reward function of the agent.

We call a shield to be an \emph{informed shield}, if it 
provides the shielded RL agent with knowledge about safety constraints.
Whenever the agent enters a state in which the shield blocks 
actions, an informed shield will send a negative reward to the learning agent for each blocked action. RL with informed shields basically explores unsafe actions  along safe execution paths instead of ignoring them completely. In contrast to unshielded RL, unsafe actions are explored for just a single step and penalised immediately.

We explore the effect of informed shields for two learning agents trained with different reward structures: one that is conservative and focuses on eating apples,  and a second one that encourages reckless behaviour that brings the agent into highly risky situations, in order to try cutting off the adversary snake. % and tries to cut off the adversary snake.

\textbf{Experimental Set-up.}
The game implementation, environment, and shield computation stay the same as in the previous experiment. The RL agent is implemented in \pytorch\cite{pytorch} and is trained for 270,000 steps with an Adam optimiser. The neural network approximating the Q-function consists of three 2-D convolution layers each followed by a batch normalisation layer. The output of the last 2-D convolution layer feeds into two stages of linear layers. 
%Before the output of the first linear layer is connected to the second one the location of the snakes and apples is appended as coordinates.
In addition to the outputs from the previous layers, the second linear layer receives the locations of the snakes and apples as additional input.
In order to reduce the complexity of the state, the input to the network is split into four channels containing information about: the map, the player snake, the enemy snake and the apples.

\subsubsection{Shielding a Conservative Agent}

In our next experiments, we considered an agent called Agent 2, trained on Map 2 (Figure~\ref{fig:training_reward_plot_agent2} (left)). The agent receives more reward for eating apples than the tabular Q-learning agent, but receives no reward for winning the game through cutting off the adversary snake. 
The resulting agent avoids getting close to the adversary snake and focuses on collecting apples. In detail, the reward function for Agent 2 is defined as follows: the agent receives +30 for eating an apple, +100 for winning by eating apples, and -100 if the game is lost or tied.
When using informed shields, the agent receives a reward of -100 from the shield for each blocked action.

\begin{figure*}[t]
    \centering
     \begin{tikzpicture}
    \node(map){
         \includegraphics[width=0.24\textwidth]{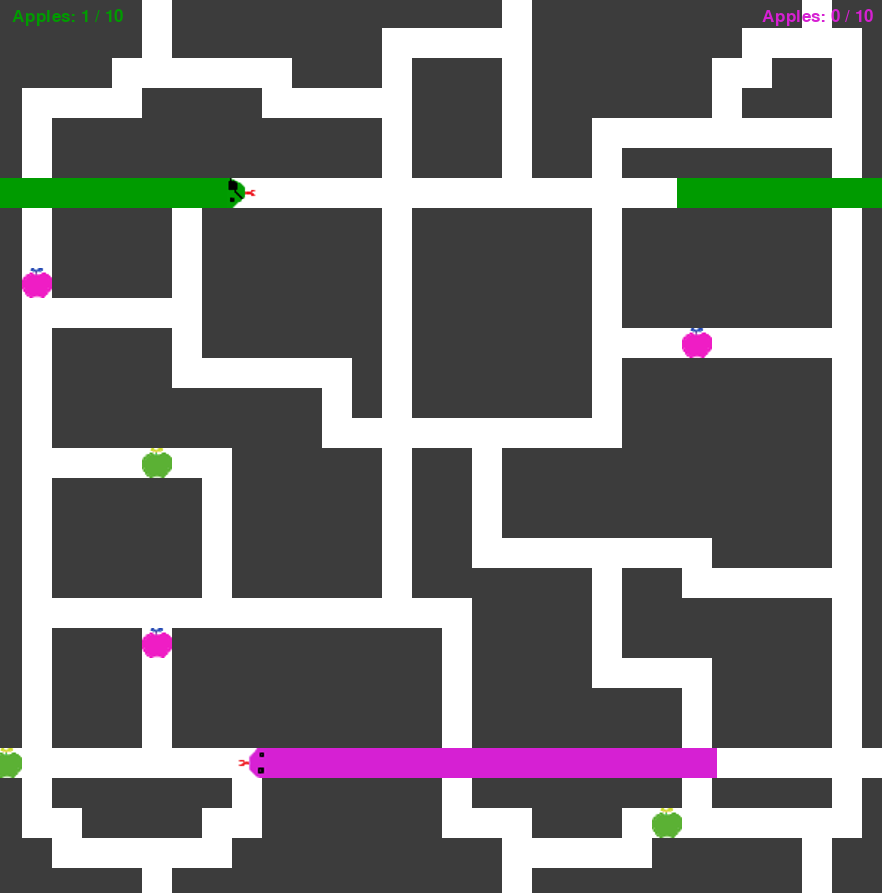}};
         \node[right = 1.2cm of map]{\includegraphics[width=0.55\textwidth]{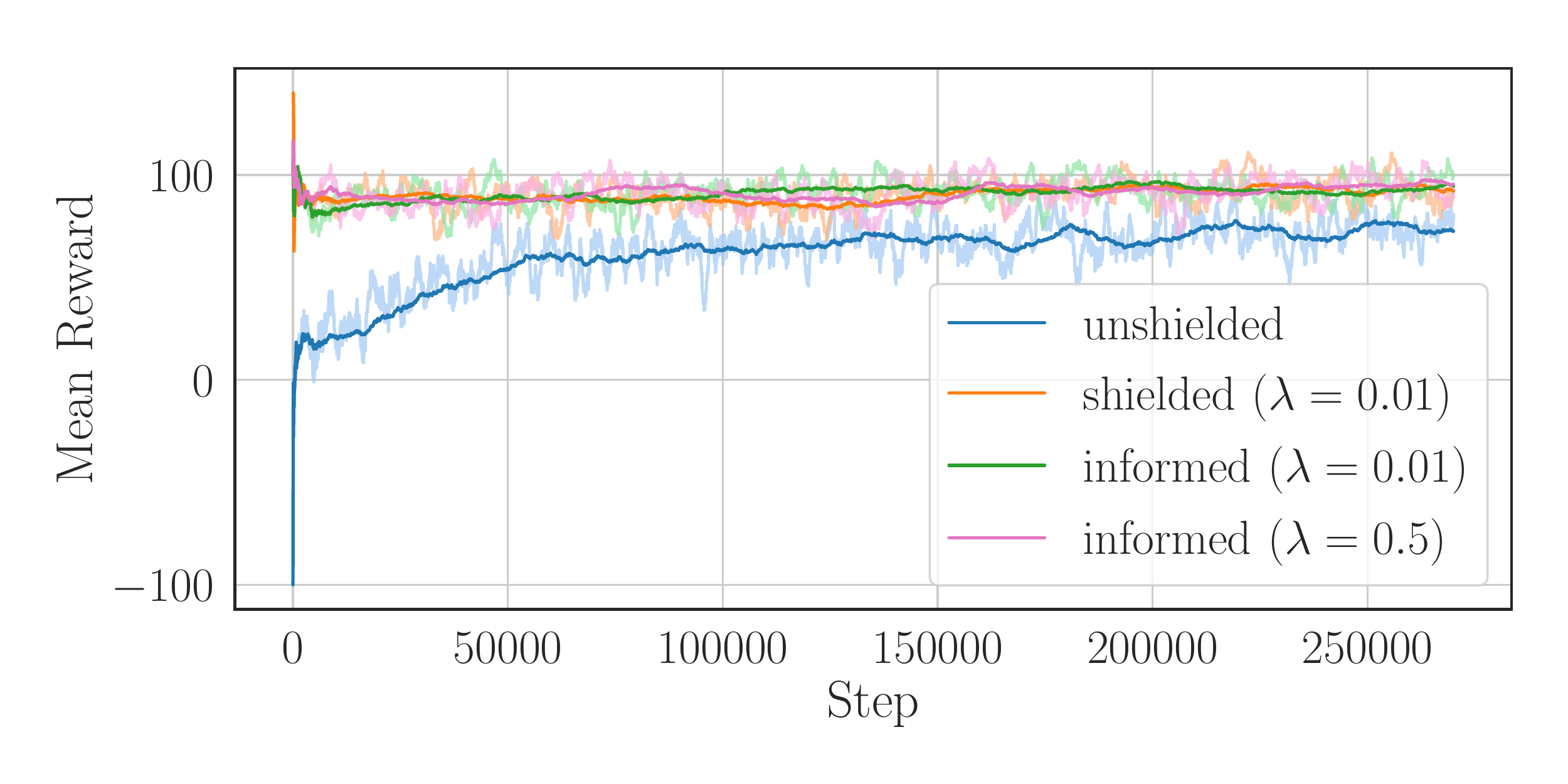}};
    \end{tikzpicture}
    \caption{Results for Agent 3. 
    Left:  A screenshot of the game using Map 3;
    Right: Reward gained in \emph{training}.}
    \label{fig:training_reward_plot}
\end{figure*}

\begin{figure*}[t]
    \centering
    \includegraphics[width=1\textwidth]{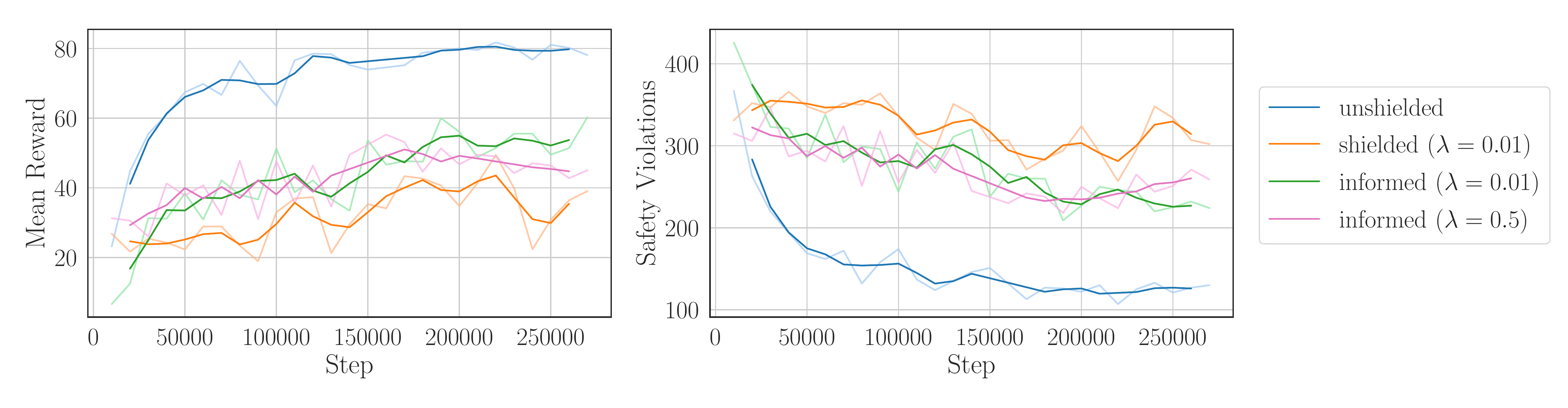}
    \caption{Intermediate execution results for Agent 3.
    Left: Reward.
    Right: Safety violations.}
    \label{fig:execution_reward_plot}
\end{figure*}

\textbf{Results.} 
\emph{Training. }
Figure~\ref{fig:training_reward_plot_agent2}~(right) shows the obtained reward for Agent 2 during training in four different settings: 
unshielded learning (blue line, no shield), shielded learning (orange line, shielded with $\lambda=0.01$), and learning with informed shields using different absolute thresholds (green line, informed shield: threshold $\lambda=0.01$; pink line, informed shield:
$\lambda=0.5$). 
The learning agents are trained for 270,000 steps. We used training steps instead of episodes (games)
for comparison, since especially in the beginning, the episodes in shielded learning are longer 
than in unshielded learning. This results from the shields preventing losing
the game.
Using the episodes in the x-axis would therefore favour the shielding approaches. The rewards at each time step are averaged over 1000 and 100 games, plotted in dark colours and light colours, respectively. 

\response{The graph shows that all shielded learning settings outperform unshielded learning.
Additionally, the informed learning settings with $\lambda=0.5$ achieved slightly better rewards than the uninformed shielding setting.
In the informed setting, the learning agent updated its Q-values
also for actions blocked by the shield with negative rewards instead 
of only performing updates for the executed actions. Even though 
the agents would not explore an unsafe action neither in the informed learning setting nor the standard shielding setting, the information about the unsafety of actions helped the learning agent to increase its learning performance.
Note that the agents shielded with $\lambda=0.01$
performed slightly worse than the agent shielded with $\lambda=0.5$.
A shield with $\lambda=0.01$ yields a very strict shield that forbids any action that is slightly risky. Thus, the shield prevents the agent
from eating apples whenever there is a slight risk of ending in a collision.
This gives the adversary snake the opportunity to be faster in eating its own apples and to win. This illustrates the trade-off between safety and performance.}

\begin{table*}[t]
    \centering
    \caption{Execution Results for Agent 3.}
    \label{tab:exe_agent3}
\scalebox{0.92}{
\begin{tabular}{|c|cc|cc|cc|cc|}
\hline
 training & \multicolumn{2}{c|}{unshielded} & \multicolumn{2}{c|}{shielded ($\lambda=0.01$)} & \multicolumn{2}{c|}{informed ($\lambda=0.01$)} & \multicolumn{2}{c|}{informed ($\lambda=0.5$)} \\
 \hline
                        & \multicolumn{1}{c|}{\AVGREW} & wins & \multicolumn{1}{c|}{\AVGREW} & wins & \multicolumn{1}{c|}{\AVGREW} & wins & \multicolumn{1}{c|}{\AVGREW} & wins \\ \hline
execution without shield    & \multicolumn{1}{c|}{78.13}   & 870     & \multicolumn{1}{c|}{39.06}   & 698     & \multicolumn{1}{c|}{60.28}   & 776     & \multicolumn{1}{c|}{45.05} & 741 \\ \hline
execution with shield & \multicolumn{1}{c|}{95.97}   &    913  & \multicolumn{1}{c|}{93.26}   &    906  & \multicolumn{1}{c|}{97.50}   &    919  & \multicolumn{1}{c|}{96.89} &  925\\ \hline
\end{tabular}}
\end{table*}

\response{\emph{Intermediate Execution.} }
To get more insights into the learned policies and the effects of shielding, we interrupt training at regular intervals to execute and evaluate the current policies.
Every 10,000 steps of training, we execute the current policy for 1000 games
\emph{without a shield}.
Figure~\ref{fig:execution_reward_plot_agent2}~(left) illustrates the obtained rewards
and Figure~\ref{fig:execution_reward_plot_agent2}~(right) shows the number of safety violations (losses of the game due to a collision). 

Since safety violations lead to a large negative reward, the values plotted on the left are negatively correlated with the values plotted on the right. 
It can be seen that policies created through unshielded learning clearly
outperform the policies obtained via shielded learning and have fewer safety violations, when the shield is removed during execution. 
As expected, the policies obtained via shielded learning without feedback about unsafe actions fail to learn safety constraints. We see the same when
using a very permissive shield that blocks actions only if the minimal probability of staying safe is less than 50 percent. 
In case of using a very strict shield that blocks all actions with a safety valuation
of greater than 0.01 (the minimal probability of reaching unsafe states), 
we observe a significant decrease in safety violations for the informed learning setting.
\response{Note that executing the agents with a shield using $\lambda=0.01$ would lead to zero safety violations.}

\subsubsection{Shielding a Reckless Agent}

In our final set of experiments, we consider an agent called Agent 3, trained on Map 3 (Figure~\ref{fig:training_reward_plot} (left)), with a reward structure that encourages the agent to cut off the enemy snake in order to win. In order to cut off the other snake, the RL agent has to get close to the adversary snake bringing itself in risky situations, which makes it interesting for shielding.
The reward structure of Agent 3 is as follows: the agent receives a reward of +10 for eating an apple, +100 for winning the game, -100 for losing, and -50 if the game ends in a tie. 
When using informed shields, the agent receives a reward of -100 from the shield for each blocked action.

\textbf{Results.}
\emph{Training.}
Figure~\ref{fig:training_reward_plot} (right) shows the rewards for Agent 3 during training. All graphs for Agent 3 use the same measurement values and metrics as for Agent 2.
\response{
The results for Agent 3 are consistent with the results obtained for Agent 2. All shielded learning settings outperformed unshielded learning. 
We observed that all three learned policies obtained with a shield did not focus on eating apples while avoiding the adversary snake, but focused on winning by
cutting off the other snake and causing it to crash into the avatar snake.}

\response{\emph{Intermediate Execution.}
Figure~\ref{fig:execution_reward_plot}~(left) illustrates the obtained rewards
and Figure~\ref{fig:execution_reward_plot}~(right) show the number of safety violations for Agent 3 during unshielded execution of policies. As before, 
every 10,000 steps of
training, we execute the current policy for 1000
games \emph{without a shield}}.

\response{We see even more significantly than before that the agent learns better to avoid safety violations when it is not shielded during training. Even in the informed case, where actions blocked by the shield are rewarded with -100,
the agent does not understand the safety objective as well as in the unshielded case. The cause for this behaviour might be that the agent receives a negative update for actions that could lead to a safety violation instead of punishing actions that immediately lead to a safety violation. This might make it more difficult for the agent to understand the safety objective, as the action that potentially causes an issue is farther from the actual 
issue.}

\response{\emph{Execution.} 
After 270.000 episodes of training, we evaluated the obtained policies. 
We executed each policy without a shield for 1000 games as well as with a shield using $\lambda=0.01$ also for 1000 games.
In Table~\ref{tab:exe_agent3} we report the averaged reward and the total number of wins. 
Using the shield in the execution results in similar results for all agents.
Since we use a shield with $\lambda=0.01$ during executions, the avatar snake only loses 
if the adversary snake manages to eat all apples and never due to a collision.
Running the agents without a shield in the execution phase shows that the 
agent that was unshielded in training outperforms all agent that were shielded in training.
}

\iffalse
\todo[inline]{Don't know if we need all the details, but here are they anyways:\\
Input 30x30 with 4 channels: map, player snake, enemy snake, apples\\
nn.Conv2d(channels, 32, kernel\_size=1, stride=1, padding=2, padding\_mode='circular') \\
nn.BatchNorm2d(32)\\
nn.Conv2d(32, 64, kernel\_size=3, stride=2)\\
nn.BatchNorm2d(64)\\
nn.Conv2d(64, 64, kernel\_size=3, stride=2)\\
nn.BatchNorm2d(64)\\
nn.Flatten()\\
nn.Linear(3136, 512)\\
nn.Linear(512+72, outputs)\\
optim.Adam(policy\_net.parameters(), lr=2e-4)\\
memory = ReplayMemory(100000)\\
BATCH\_SIZE = 512\\
GAMMA = 0.9\\
EPS\_START = 0.9\\
EPS\_END = 0.1\\
EPS\_DECAY = 18000}
\fi

\section{Conclusion and Future Work}\label{sec:conclusion}

\response{In this paper, we propose an approach to prevent safety violations that can be avoided by planning ahead a short time into the future. }  
Our online shielding exploits the time required to complete tasks to model and analyse the immediate future with respect to a safety property. For every decision at runtime, we create MDPs to model the current state of the environment and the behaviour of the agents. Given these MDP models, we employ probabilistic model-checking to evaluate every action possible in the next decision. In particular, we determine the probability of unsafe behaviours following every possible choice. This information is used to block unsafe actions, i.e., actions leading to safety violations with a probability exceeding a threshold relative to the minimal probability of safety violations.

We evaluate online shielding in the context of RL, by empirically analysing the effect of shielding on learning performance and the safety of learned policies. For this purpose, we proposed informed shields that update the learner's value function by penalising blocked actions and compare unshielded RL and RL augmented with uninformed and informed shields. Our experimental results show that shielding improves the performance of RL agents during learning. However, the final learned policies inflict more safety violations than conventionally learned policies, when executed in unshielded environments. Hence, to guarantee safety of control policies obtained through shielded (or unshielded) RL, shielding needs to be applied during execution in the field.

% For future work, we plan to investigate the application of online shielding in other settings, such as decision making in robotics and control. 
For future work, we plan to investigate the influence of imperfect information on shielding and shielded reinforcement learning. Online shielding is well-suited for agents using unreliable sensors, as it could counter sensor defects at runtime as they occur. 
We also plan to study the application of online shielding in other settings, such as decision making in robotics and control. Another interesting extension would be to incorporate quantitative performance measures in the form of rewards and costs into the computation of the online shield, as previously demonstrated in an offline manner~\cite{DBLP:conf/cav/AvniBCHKP19} and in a hybrid approach~\cite{DBLP:journals/corr/abs-2010-03842}, where runtime information was used to learn the environment dynamics.

\subsubsection*{Acknowledgments.}
This work has been supported by the "University SAL Labs" initiative of Silicon Austria Labs (SAL) and its Austrian partner universities for applied fundamental research for electronic based systems.
Additionally, this project has received funding from the European Union’s Horizon 2020 research
and innovation programme under grant agreement N◦ 956123 - FOCETA.

\bibliographystyle{sn-basic}
\bibliography{literature}

\end{document}